\documentclass[runningheads]{llncs}
\usepackage{graphicx}

\usepackage{tikz}
\usepackage{comment}
\usepackage{amsmath,amssymb} 
\usepackage{color}

\usepackage[accsupp]{axessibility}  

\usepackage[pagebackref,breaklinks,colorlinks]{hyperref}

\usepackage{graphicx}
\usepackage{amsmath}
\usepackage{amssymb}
\usepackage{booktabs}
\usepackage{subfigure}
\usepackage{multirow}
\usepackage{bm}
\usepackage{pifont}

\newcommand{\ie}{\emph{i.e., }}
\newcommand{\eg}{\emph{e.g., }}
\renewcommand{\paragraph}{\subsubsection}

\usepackage[capitalize]{cleveref}
\crefname{section}{Sec.}{Secs.}
\Crefname{section}{Section}{Sections}
\Crefname{table}{Table}{Tables}
\crefname{table}{Tab.}{Tabs.}

\newcommand{\cmark}{\ding{51}}
\newcommand{\xmark}{\ding{55}}

\newcommand{\cL}{\mathcal{L}}

\newcommand{\cD}{\mathcal{D}}
\newcommand{\cM}{\mathcal{M}}
\newcommand{\R}{\mathbb{R}}

\begin{document}
\pagestyle{headings}
\mainmatter
\def\ECCVSubNumber{4388}  

\title{Optical Flow Training under Limited Label Budget via Active Learning} 

\titlerunning{Optical Flow Training under Limited Label Budget via Active Learning}
%
\author{
Shuai Yuan\and
Xian Sun\and
Hannah Kim \and
Shuzhi Yu \and
Carlo Tomasi
}
\authorrunning{S. Yuan et al.}
%
\institute{
Duke University, Durham NC 27708, USA\\
\email{\{shuai,hannah,shuzhiyu,tomasi\}@cs.duke.edu, xian.sun@duke.edu}
}
\maketitle

\begin{abstract}
Supervised training of optical flow predictors generally yields better accuracy than unsupervised training. However, the improved performance comes at an often high annotation cost. Semi-supervised training trades off accuracy against annotation cost. We use a simple yet effective semi-supervised training method to show that even a small fraction of labels can improve flow accuracy by a significant margin over unsupervised training. In addition, we propose active learning methods based on simple heuristics to further reduce the number of labels required to achieve the same target accuracy. Our experiments on both synthetic and real optical flow datasets show that our semi-supervised networks generally need around 50\% of the labels to achieve close to full-label accuracy, and only around 20\% with active learning on Sintel. We also analyze and show insights on the factors that may influence active learning performance. Code is available at \url{https://github.com/duke-vision/optical-flow-active-learning-release}.
\end{abstract}


\section{Introduction} \label{sec:introduction}


The estimation of optical flow is a very important but challenging task in computer vision with broad applications including video understanding~\cite{fan2018end}, video editing~\cite{gao2020flow}, object tracking~\cite{aslani2013optical}, and autonomous driving~\cite{kitti15}.

Inspired by the successes of deep CNNs in various computer vision tasks~\cite{krizhevsky2012imagenet,he2016deep}, much recent work has modeled optical flow estimation in the framework of supervised learning, and has proposed several networks of increasingly high performance on benchmark datasets~\cite{dosovitskiy2015flownet,ranjan2017optical,sun2018pwc,hui2018liteflownet,hur2019iterative,teed2020raft,zhang2021separable}. Ground-truth labels provide a strong supervision signal when training these networks. However, ground-truth optical flow annotations are especially hard and expensive to obtain. Thus, many methods use synthetic data in training, since ground-truth labels can be generated as part of data synthesis. Nevertheless, it is still an open question whether synthetic data are an adequate proxy for real data.

Another way to circumvent label scarcity is unsupervised training, which does not require any labels at all. Instead, it relies on unsupervised loss measures that enforce exact or approximate constraints that correct outputs should satisfy. Common losses used in unsupervised optical flow estimation are the photometric loss, which penalizes large color differences between corresponding points, and the smoothness loss, which penalizes abrupt spatial changes in the flow field~\cite{jason2016back,ren2017unsupervised,meister2018unflow,janai2018unsupervised,liu2019selflow,liu2020learning}. While unsupervised methods allow training on large datasets from the application domain, their performance is still far from ideal because the assumed constraints do not always hold. For instance, the photometric loss works poorly with non-Lambertian surfaces or in occlusion regions~\cite{wang2018occlusion}, while the smoothness loss fails near motion discontinuities~\cite{kim}.

\begin{figure*}[t]
  \centering
    \includegraphics[width=\linewidth]{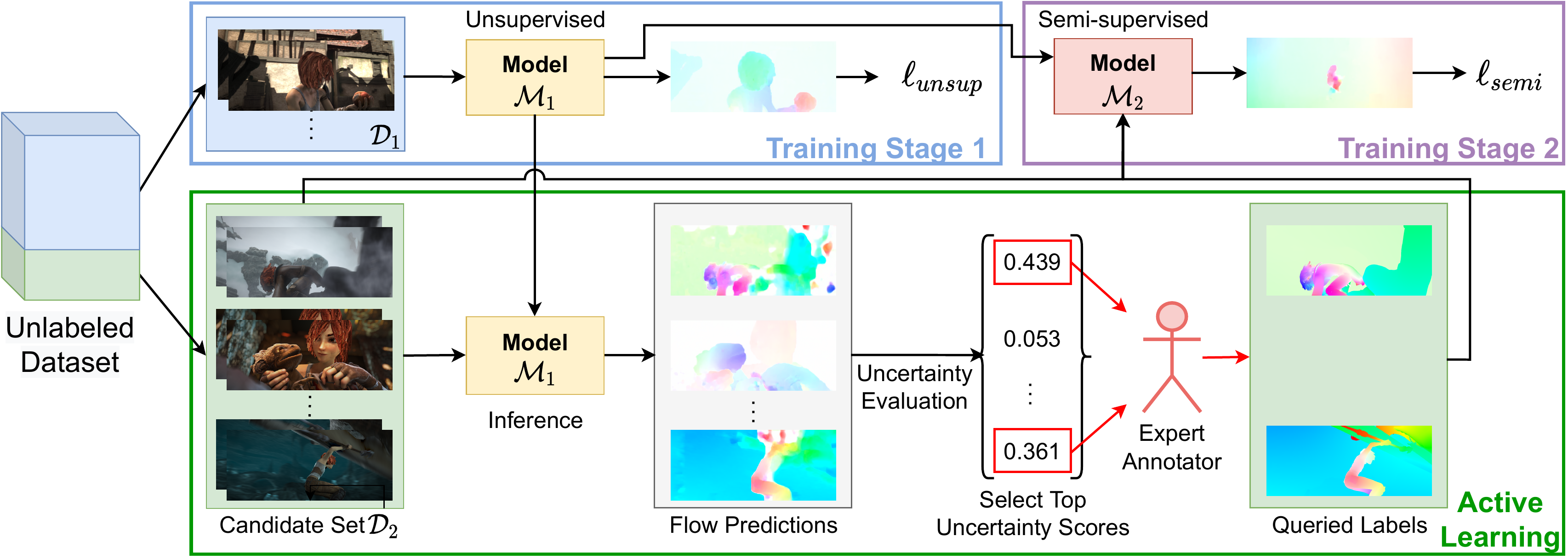}
  \caption{Overview of our active learning framework for the semi-supervised training.}
  \label{fig:overview}
\end{figure*}

Semi-supervised training can be a way to combine the advantages of both supervised and unsupervised training for optical flow models. The idea is simple, and amounts to training the network with a mix of labeled and unlabeled data. This is possible because we can charge different losses (supervised or unsupervised) to different samples depending on whether they are labeled or not.

The trade-off between performance and labeling cost is of interest in real practice, since it describes the marginal benefit that can be accrued at the price of a unit of labeling effort. However, little work has focused on the semi-supervised training of optical flow. Existing methods have tried to improve flow estimates given an available, partially labeled dataset~\cite{yang2018conditional,lai2017semi,song2020fpcr}. Other work uses semi-supervised training to address specific problem conditions, \eg foggy scenes~\cite{yan2020optical}.

In contrast, we are particularly interested in label efficiency, that is, in the performance improvement gained as the fraction of labeled samples increases from 0 (``unsupervised'') to 1 (``supervised''). Specifically, we use a simple yet effective semi-supervised algorithm and show that the model error drops significantly as soon as a small fraction of the samples are labeled. This suggests that even a modest labeling budget can lead to a significant performance boost.

Given a specific labeling budget, an important related question is how to determine which part of the dataset to label. A simple method is random sampling, but it is possible to do better. Specifically, we propose and evaluate criteria that suggest whose labels bring larger benefits in training. This brings us to the concept of active learning.

Active Learning (AL) has been shown to be effective in reducing annotation costs while maintaining good performance in many vision tasks including image classification~\cite{li2013adaptive,beluch2018power}, object detection~\cite{roy2018deep,choi2021active}, semantic segmentation~\cite{mackowiak2018cereals,siddiqui2020viewal}, and instance segmentation~\cite{wang2017incorporating}. The general idea is to allow the training algorithm to select valuable unlabeled samples for which to query labels for further training. This selection is especially important for optical flow estimation, since generating labels for additional samples incurs high costs in terms of computation, curation, and sometimes even hand annotations.

While annotating individual flow vectors by hand is effectively impossible in practice, annotation can be and often is done by hand at a higher level and, even so, is costly. For instance, in KITTI 2015~\cite{kitti15}, correspondences between points on CAD models of moving cars are annotated by hand so that dense optical flow can be inferred for these cars. In addition, nonrigid objects such as pedestrians or bicyclists are manually masked out, and so are errors in the flow and disparity masks inferred from LiDAR and GPS/IMU measurements and from stereo depth estimation. This is still manual annotation and curation, painstaking and expensive. Some amount of curation, at the very least, is necessary for most high-quality training sets with real imagery, and the methods we propose aim to reduce the need for this type of work, and to make the products of whatever manual work is left more effective.
To the best of our knowledge, we are the first to study active learning as a way to moderate the high annotation costs for optical flow estimation.

As illustrated in \cref{fig:overview}, our training pipeline (top part of the diagram) includes an unsupervised first stage and a semi-supervised second stage. We split our unlabeled dataset to two sets, one ($\cD_1$) used to pre-train an unsupervised model $\cM_1$ and the other ($\cD_2$) used as the \emph{candidate} set, from which samples are selected to query labels from expert annotators. After training model $\cM_1$ on $\cD_1$ in Stage 1, we estimate flow for all the samples in $\cD_2$ and score each of them based on our active learning criteria. We query for labels for top-scoring samples and add these to $\cD_2$ for further semi-supervised training in Stage 2. In this paper, we show that using active learning to query labels can help further reduce the number of labels required to achieve a given performance target in semi-supervised training.

In summary, our contributions are as follows.
\begin{itemize}
    \item We show on several synthetic and real-life datasets that the performance from  unsupervised training of optical flow estimators can be improved significantly as soon as a relatively small fraction of labels are added for semi-supervised training.
    \item To the best of our knowledge, we are the first to explore active learning as a way to save annotation cost for optical flow estimation, and our novel pipeline can be used directly in real practice.
    \item We set up the new problem of semi-supervised training of optical flow under certain label ratio constraints. We anticipate follow-up research to propose better methods for this problem.
\end{itemize}



\section{Related Work} \label{sec:related_work}

\textbf{Supervised Optical Flow}~ 
Supervised methods use deep networks to learn the mapping from image pairs to the corresponding optical flow by minimizing the supervised loss, namely, some distance measure between computed and true flow. FlowNet~\cite{dosovitskiy2015flownet} used a multi-scale encoder-decoder structure with skip connections between same-scale layers. Following this framework, many networks have been proposed to decrease both model size and error. Traditional ideas or heuristics have been introduced into the network, including image pyramid in SPyNet~\cite{ranjan2017optical}, feature pyramid, warping, and cost volume in PWC-Net~\cite{sun2018pwc} and LiteFlowNet~\cite{hui2018liteflownet}. Iterative decoder modules have also been explored as a way to reduce model size while retaining accuracy in IRR-PWC~\cite{hur2019iterative} and RAFT~\cite{teed2020raft}. The latter built the network based on full-pair correlations and has led to many follow-up models that have achieved the state-of-the-art performance~\cite{zhang2021separable}.


\textbf{Unsupervised Optical Flow}~ 
Recent research has focused on the unsupervised learning of optical flow as a compromise between label availability and model performance. Initial work on this topic proposed to train FlowNet-like networks using surrogate loss terms, namely photometric loss and smoothness loss~\cite{jason2016back,ren2017unsupervised}. As found by many papers, flow at occlusion region is especially challenging for unsupervised networks~\cite{wang2018occlusion}. Thus, much research focused on solving the occlusion problem via occlusion masks~\cite{wang2018occlusion}, bi-directional consistency~\cite{meister2018unflow}, multi-frame consistency~\cite{janai2018unsupervised,ren2019fusion}, and self-supervised teacher-student models~\cite{liu2019selflow,yu2020motion}. ARFlow~\cite{liu2020learning} integrated a second forward pass using transformed inputs for augmentation and has achieved the state-of-the-art unsupervised performance. Multi-frame unsupervised models have also been investigated~\cite{janai2018unsupervised,stone2021smurf}.

\textbf{Semi-supervised Training in Vision}~
Semi-supervised training targets applications where partial labels are available. Early approaches in image classification~\cite{laine,sajjadi,miyato,grandvalet} utilize label propagation with regularization and augmentation based on the belief that nearby data points tend to have similar class labels. A more recent class of methods train on unlabeled samples with pseudo-labels~\cite{lee,xie} predicted by a supervised trained network trained with labeled samples. Similar teacher-student models have also been explored~\cite{liu2019structured,tarvainen2017mean}.

Although widely explored in many other vision tasks, there is little work on semi-supervised optical flow. Some early work utilized semi-supervised learning to achieve comparable flow accuracy to the supervised methods~\cite{lai2017semi,song2020fpcr,yang2018conditional,lai2017semi}. Others applied semi-supervised methods to tackle specific cases of optical flow, such as dense foggy scenes\cite{yan2020optical} and ultrasound elastography\cite{tehrani2020semi}. In contrast, we focus on label efficiency for optical flow estimation: Instead of proposing semi-supervised networks that focus on improving benchmark performances by adding external unlabeled data, we are more focused on the trade-off between performance and label ratio given a fixed dataset. 

\textbf{Active Learning in Vision}~ 
 Active Learning (AL) aims to maximize model performance with the least amount of labeled data by keeping a human in the training loop. The general idea is to make the model actively select the most valuable unlabeled samples and query the human for labels which are used in the next stage of training. There are two main categories, namely, uncertainty-based (select samples based on some pre-defined uncertainty metric)~\cite{kapoor2007active,houlsby2011bayesian,gal2017deep,ebrahimi2019uncertainty}, and distribution-based (query representative samples of sufficient diversity)~\cite{paul2017non,wei2015submodularity}. 

Active learning has achieved extensive success in various fields in computer vision, including image classification~\cite{li2013adaptive,beluch2018power}, object detection~\cite{roy2018deep,choi2021active}, semantic segmentation~\cite{mackowiak2018cereals,siddiqui2020viewal}, and instance segmentation~\cite{wang2017incorporating}. However, the concept has received little attention in optical flow estimation where acquiring labels is especially difficult. To the best of our knowledge, we are the first to apply active learning to optical flow estimation to reduce annotation cost.

\section{Method} \label{sec:method}

As we are among the first to explore active learning as a way to tackle the high annotation costs in optical flow training, we start from simple yet effective methods to implement our ideas. This section describes our semi-supervised training method (\cref{sec:semi-sup_train}), active learning heuristics (\cref{sec:active_heuristics}), and network structure and loss functions (\cref{sec:network_and_loss}).

\subsection{Semi-supervised Training}\label{sec:semi-sup_train}


Given a partially labeled data set, we implement the semi-supervised training by charging a supervised loss to the labeled samples and an unsupervised loss to the unlabeled ones. Specifically, the semi-supervised loss for each sample $\bm x$ is
\begin{equation} \label{eq:semi-sup_loss}
\ell_{\text{semi}}(\bm x) = \left\{\begin{array}{ll}
\ell_{\text{unsup}}(\bm x),   & \text{if $\bm x$ is unlabeled,} \\
\alpha \ell_{\text{sup}}(\bm x),   & \text{otherwise}
\end{array}\right.
\end{equation}
where $\alpha>0$ is a balancing weight. We do not include the unsupervised loss for labeled samples (although in principle this is also an option) to avoid any conflict between the two losses, especially on occlusion and motion boundary regions.

Thus, the final loss of the data set $\cD=\cD^u\cup\cD^l$ is
\begin{equation}
    \cL_{\text{semi}} = \sum_{\bm x\in \cD} \ell_{\text{semi}}(\bm x) = \sum_{\bm x\in \cD^{u}} \ell_{\text{unsup}}(\bm x) + \alpha \sum_{\bm x\in \cD^{l}} \ell_{\text{sup}}(\bm x),
\end{equation}
where $\cD^u$ and $\cD^l$ are the unlabeled and labeled sets. We define the \emph{label ratio} as $r=|\cD^l|/|\cD|$. During training, we randomly shuffle the training set $\cD$, so that each batch of data has a mix of labeled and unlabeled samples.

\subsection{Active Learning Heuristics}\label{sec:active_heuristics}

\Cref{fig:overview} shows a general overview of our active learning framework. After pre-training our model on unlabeled data (Stage 1), we invoke an active learning algorithm to determine samples to be labeled for further training. Specifically, we first use the pre-trained model to infer flow on the samples of another disjoint unlabeled data set (the \emph{candidate} set) and select a fraction of the samples to be labeled, based on some criterion. After obtaining those labels, we continue to train the model on the partially labeled candidate set using the semi-supervised loss (Stage 2). Note that in this second stage, we do not include the unlabeled data used in pre-training (see ablation study in \cref{sec:main_results}). By allowing the model to actively select samples to query labels, we expect the model to achieve the best possible performance under a fixed ratio of label queries (the ``label budget'').

So, what criteria should be used for selecting samples to be labeled? Many so-called uncertainty-based methods for active learning algorithms for image classification or segmentation use the soft-max scores to compute how confident the model is about a particular output. However, optical flow estimation is a regression problem, not a classification problem, so soft-max scores are typically not available, and would be in any case difficult to calibrate.

Instead, we select samples for labeling based on heuristics specific to the optical flow problem. For example, the photometric loss is low for good predictions. In addition, unsupervised flow estimation performs poorly at occlusion regions and motion discontinuities. These considerations suggest the following heuristic metrics to flag points for which unsupervised estimates of flow are poor:
\begin{itemize}
    \item \emph{Photo loss}: the photometric loss used in training.
    \item \emph{Occ ratio}: the ratio of occlusion pixels in the frame, with occlusion estimated by consistency check of forward and backward flows~\cite{meister2018unflow}.
    \item \emph{Flow grad norm}: the magnitude of gradients of the estimated flow field as in \cite{ilg2018occlusions} averaged across the frame, used to indicate the presence of motion boundaries.
\end{itemize}
We experiment with three active learning methods, each using one of the metrics above. When querying labels for a given label ratio $r$, we first compute the metric for each sample in the candidate set, and then sort and pick the samples with largest uncertainties as our queries.

\subsection{Network Structure and Loss Functions}\label{sec:network_and_loss}

\paragraph{Network Structure} We adopt the unsupervised state-of-the-art, ARFlow~\cite{liu2020learning}, as our base network, which is basically a lightweight variant of PWC-Net~\cite{sun2018pwc}. PWC-Net-based structures have been shown to be successful in both supervised and unsupervised settings, so it is a good fit for our hybrid semi-supervised training. We do not choose RAFT because it has been mostly proven to work well in the supervised setting, while our setting (\cref{sec:experimental_settings_active}) is much closer to the unsupervised one (see appendix for details).

Each sample is a triple $\bm x=(I_1, I_2, U_{12})$ where $I_1, I_2\in\R^{h\times w\times 3}$ are the two input frames and $U_{12}$ is the true optical flow (set as ``None'' for unlabeled samples). The network estimates a multi-scale forward flow field $f(I_1, I_2)=\{\hat{U}_{12}^{(2)}, \hat U_{12}^{(3)}, \cdots, \hat U_{12}^{(6)}\}$, where the output $\hat U_{12}^{(l)}$ at scale $l$ has dimension $\frac{h}{2^l}\times\frac{w}{2^l}\times 2$. The finest estimated scale is $\hat U_{12}^{(2)}$, which is up-sampled to yield the final output.

\paragraph{Unsupervised Loss} 
For unsupervised loss $\ell_{\text{unsup}}(\bm x)$ we follow ARFlow\cite{liu2020learning}, which includes a photometric loss $\ell_{\text{ph}}(\bm x)$, a smoothness loss $\ell_{\text{sm}}(\bm x)$, and an augmentation loss $\ell_{\text{aug}}(\bm x)$:
\begin{equation} \label{eq:unsup_loss}
    \ell_{\text{unsup}}(\bm x) = \ell_{\text{ph}}(\bm x) + \lambda_{\text{sm}} \ell_{\text{sm}}(\bm x) + \lambda_{\text{aug}} \ell_{\text{aug}}(\bm x).
\end{equation}

Specifically, given the sample $\bm x$, we first estimate both forward and backward flow, $\hat U_{12}^{(l)}$ and $\hat U_{21}^{(l)}$, and then apply forward-backward consistency check~\cite{meister2018unflow} to estimate their corresponding occlusion masks, $\hat O_{12}^{(l)}$ and $\hat O_{21}^{(l)}$.

To compute the photometric loss, we first warp the frames by $\hat I_1^{(l)}(\bm p) = I_2^{(l)}(\bm p + \hat U_{12}^{(l)}(\bm p))$, where $I_2^{(l)}$ is $I_2$ down-sampled to the $l$-th scale and $\bm p$ denotes pixel coordinates at that scale. The occlusion-aware photometric loss at each scale can be then defined as
\begin{equation} \label{eq:ph_loss_one_scale}
    \ell_{\text{ph}}^{(l)}(\bm x) = \sum_{i=1}^3 c_i \ \rho_i(\hat I_1^{(l)}, I_1^{(l)}, \hat O_{12}^{(l)})
\end{equation}
where $\rho_1, \rho_2, \rho_3$ are three distance measures with the estimated occlusion region filtered out in computation. As proposed in~\cite{liu2020learning}, these three measures are the L$_1$-norm, structural similarity (SSIM)~\cite{wang2004image}, and the ternary census loss~\cite{meister2018unflow}, respectively, weighted by $c_i$.

The edge-aware smoothness loss of each scale $l$ is computed using the second-order derivatives:
\begin{equation}
    \ell_{\text{sm}}^{(l)}(\bm x) = \frac{1}{2|\Omega^{(l)}|} \sum_{z\in\{x, y\}}\sum_{\bm p\in\Omega^{(l)}} \left\|\frac{\partial^2 \hat U_{12}^{(l)}(\bm p)}{\partial z^2}\right\|_1 e^{-\delta\left\|\frac{\partial I_1(\bm p)}{\partial z}\right\|_1},
\end{equation}
where $\delta=10$ is a scaling parameter, and $\Omega^{(l)}$ denotes the set of pixel coordinates on the $l$-th scale.

We combine the losses of each scale linearly using weights $w_{\text{ph}}^{(l)}$ and $w_{\text{sm}}^{(l)}$ by
\begin{equation} \label{eq:ph_sm_loss}
\ell_{\text{ph}}(\bm x) = \sum_{l=2}^6w_{\text{ph}}^{(l)}\ell_{\text{ph}}^{(l)}(\bm x),\quad\ell_{\text{sm}}(\bm x) = \sum_{l=2}^6 w_{\text{sm}}^{(l)}\ell_{\text{sm}}^{(l)}(\bm x).
\end{equation}
We also include the photometric and smoothness loss for the backward temporal direction, which is not shown here for conciseness.

After the first forward pass of the network, ARFlow also conducts an additional forward pass on input images transformed with random spatial, appearance, and occlusion transformations to mimic online augmentation. The augmentation loss $\ell_{\text{aug}}(\bm x)$ is then computed based on the consistency between outputs before and after the transformation. See~\cite{liu2020learning} for details.

\paragraph{Supervised Loss} 
For supervised loss $\ell_{\text{sup}}(\bm x)$, we apply the multi-scale robust L$_1$-norm
\begin{equation} \label{eq:sup_loss}
    \ell_{\text{sup}}(\bm x) = \sum_{l=2}^6 \frac{w_{\text{sup}}^{(l)}}{|\Omega^{(l)}|}\sum_{\bm p\in\Omega^{(l)}}(\|\hat U_{12}^{(l)}(\bm p) - U_{12}^{(l)}(\bm p)\|_1 + \epsilon)^q,
\end{equation}
where $U_{12}^{(l)}$ is the down-sampled true flow to the $l$-th scale. A small $\epsilon$ and $q<1$ is included to penalize less on outliers. We set $\epsilon=0.01$ and $q=0.4$ as in~\cite{sun2018pwc}.

\paragraph{Semi-supervised Loss}
The semi-supervised loss is computed by \cref{eq:semi-sup_loss}. 

\section{Experimental Results} \label{sec:experimental_results}

\subsection{Datasets} 
As most optical flow methods, we train and evaluate our method on FlyingChairs~\cite{dosovitskiy2015flownet}, FlyingThings3D~\cite{flyingthings3D}, Sintel~\cite{sintel}, and KITTI~\cite{kitti12,kitti15} datasets. 
Apart from the labeled datasets, raw Sintel and KITTI frames with no labels are also available and often used in recent unsupervised work~\cite{liu2019selflow,liu2020learning,ranjan2019competitive,yin2018geonet}. As common practice, we have excluded the labeled samples from the raw Sintel and KITTI datasets. 


In our experiments, we also split our own train and validation set on Sintel and KITTI. We split Sintel clean and final passes by scenes to 1,082 training samples and 1,000 validation samples. For KITTI, we put the first 150 samples in each of 2015 and 2012 set as our training set, yielding 300 training samples and 94 validation samples. A summary of our data splits is in the appendix.

\subsection{Implementation Details}\label{sec:implementation_details} 

We implement the model in PyTorch~\cite{NEURIPS2019_9015}, and all experiments share the same hyper-parameters as follows. Training uses the Adam optimizer~\cite{kingma2014adam} with $\beta_1=0.9$, $\beta_2=0.999$ and batch size 8. The balancing weight $\alpha$ in \cref{eq:semi-sup_loss} is set as 1. The weights of each unsupervised loss term in \cref{eq:unsup_loss} are $\lambda_{\text{sm}} = 50$ for Sintel and $\lambda_{\text{sm}} = 75$ otherwise; and $\lambda_{\text{aug}}=0.2$ unless otherwise stated. The weights of different distance measures in \cref{eq:ph_loss_one_scale} are set as $(c_1, c_2, c_3)=(0.15, 0.85, 0)$ in the first 50k iterations and $(c_1, c_2, c_3)=(0, 0, 1)$ in the rest as in ARFlow~\cite{liu2020learning}.

The supervised weights $w_{\text{sup}}^{(l)}$ for scales $l=2,3,\cdots,6$ in \cref{eq:sup_loss} are 0.32, 0.08, 0.02, 0.01, 0.005 as in PWC-Net\cite{sun2018pwc}. The photometric weights $w_{\text{ph}}^{(l)}$ in \cref{eq:ph_sm_loss} are 1, 1, 1, 1, 0, and the smoothness weights $w_{\text{sm}}^{(l)}$ in \cref{eq:ph_sm_loss} are 1, 0, 0, 0, 0.

For data augmentation, we include random cropping, random rescaling, horizontal flipping, and appearance transformations (brightness, contrast, saturation, hue, Gaussian blur). Please refer to the appendix for more details.

\subsection{Semi-supervised Training Settings}\label{sec:experimental_settings_semi-sup}

The goal of this first experiment is to see how the validation error changes as we gradually increase the label ratio $r$ from 0 (unsupervised) to 1 (supervised). We are specifically interested in the changing error rate, which reflects the marginal gain of a unit of labeling effort.

We ensure that all experiments on the same dataset have exactly the same setting except the label ratio $r$ for fair comparison. For each experiment, the labeled set is sampled uniformly. We experiment on all four datasets independently using label ratio $r\in\{0, 0.05, 0.1, 0.2, 0.4, 0.6, 0.8, 1\}$ with settings below.

\paragraph{FlyingChairs and FlyingThings3D} As a simple toy experiment, we split the labeled and unlabeled sets randomly and train using the semi-supervised loss. We train for 1,000k iterations with a fixed learning rate $\eta=0.0001$.

\paragraph{Sintel} Unlike the two large datasets above, Sintel only has ground-truth labels for 2,082 clean and final samples, which is too small to train a flow model effectively on its own. Thus, the single-stage schedule above may not apply well.

Instead, as is common practice in many unsupervised methods, we first pre-train the network using the large Sintel raw movie set in an unsupervised way. Subsequently, as the second stage, we apply semi-supervised training with different label ratios on our training split of clean and final samples. Note that we compute the label ratio $r$ as the ratio of labeled samples only in our second-stage train split, which does not include the unlabeled raw data samples in the first stage. This is because the label ratio would otherwise become too small (thus less informative) since the number of raw data far exceeds clean and final data.

We train the first stage using learning rate $\eta=0.0001$ for 500k iterations, while the second stage starts with $\eta=0.0001$, which is cut by half at 400, 600, and 800 epochs, and ends at 1,000 epochs. Following ARFlow~\cite{liu2020learning}, we turn off the augmentation loss by assigning $\lambda_{\text{aug}}=0$ in the first stage.

\paragraph{KITTI} We apply a similar two-stage schedule to KITTI. We first pre-train the network using KITTI raw sequences with unsupervised loss. Subsequently, we assign labels to our train split of the KITTI 2015/2012 set with a given label ratio by random sampling and then run the semi-supervised training. The learning rate schedule is the same as that for Sintel above.

\subsection{Active Learning Settings}\label{sec:experimental_settings_active}

The second part of experiments is on active learning, where we show that allowing the model to select which samples to label can help reduce the error.

We mainly experiment on Sintel and KITTI since they are close to real data. Since active learning is a multi-stage process (which needs a pre-trained model to query labels for the next stages), it fits well with the two-stage semi-supervised settings described in \cref{sec:experimental_settings_semi-sup}. Thus, we use those settings with labels queried totally at random as our baseline. In comparison, we show that using the three active learning heuristics described in \cref{sec:active_heuristics} to query labels can yield better results than random sampling. We try small label ratios $r\in\{0.05, 0.1, 0.2\}$ since the semi-supervised training performance starts to saturate at larger label ratios.

\subsection{Main Results} \label{sec:main_results} 

\begin{figure}[tbp] 
  \centering
  \subfigure[FlyingChairs]{
     \includegraphics[width=0.27\linewidth]{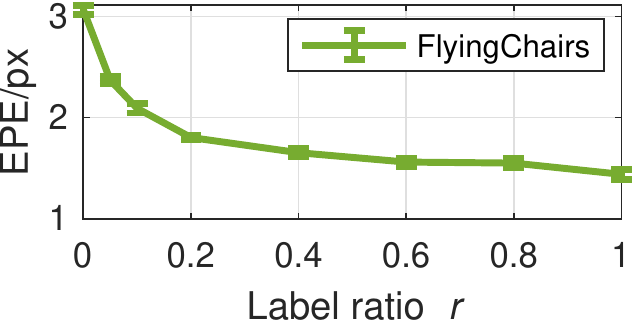} 
    \label{fig:chairs_curve_error_bar}
  }
  \subfigure[Sintel]{
    \includegraphics[width=0.51\linewidth]{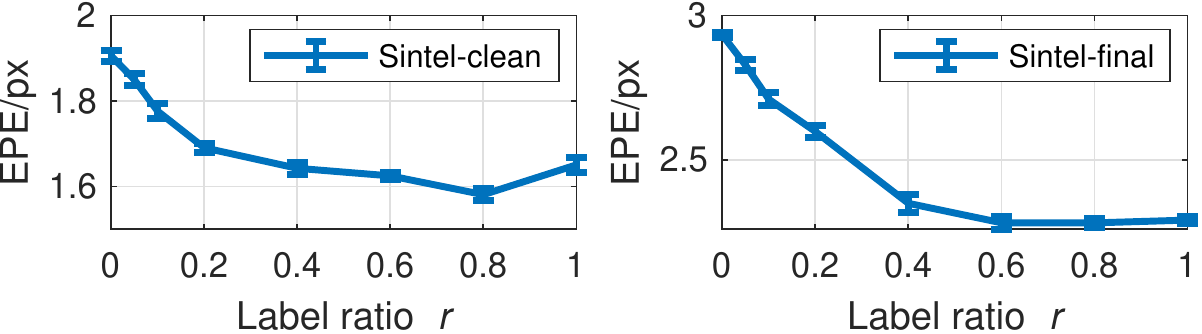} 
    \label{fig:sintel_curve_error_bar}
  }  
  \subfigure[FlyingThings]{
    \includegraphics[width=0.27\linewidth]{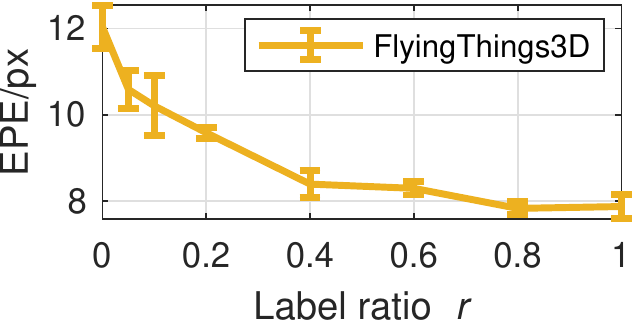} 
    \label{fig:things3d_curve_error_bar}
  }
  \subfigure[KITTI]{
    \includegraphics[width=0.51\linewidth]{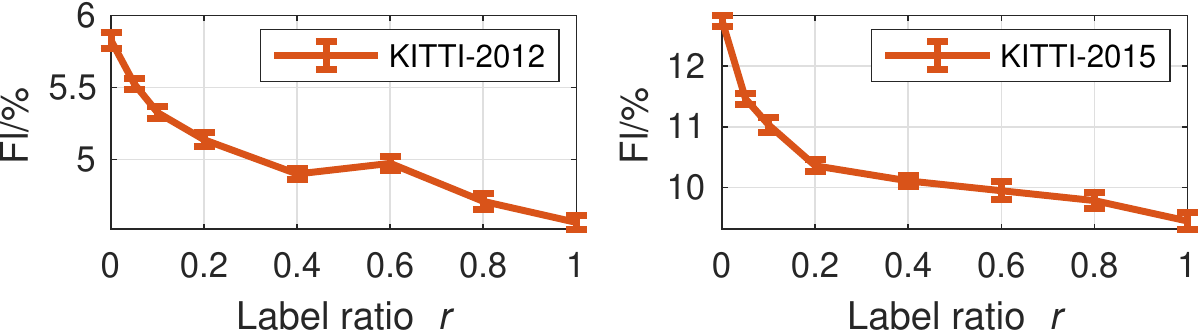} 
    \label{fig:kitti_curve_error_bar}
  }
  \caption{Model validation errors of the semi-supervised training with different label ratios. `EPE': End-Point Error, `Fl': Flow error percentage.}
  \label{fig:exp1}
\end{figure}

\paragraph{Semi-supervised Training} We first experiment with the semi-supervised training with different label ratios across four commonly used flow datasets. As shown in \cref{fig:exp1}, the model validation error drops significantly at low label ratios and tends to saturate once an adequate amount of labels are used. This supports our hypothesis that even a few labels can help improve performance significantly.

Another observation is that the errors for FlyingChairs, FlyingThings3D, and Sintel saturate at around 50\% labeling, whereas KITTI keeps improving slowly at high label ratios. One explanation for this discrepancy may involve the amount of repetitive information in the dataset: Sintel consists of video sequences with 20-50 frames that are very similar to each other, while KITTI consists of individually-selected frame pairs independent from the other pairs.

\paragraph{Active Learning}

\begin{figure}[tbp] 
  \centering
  \subfigure[Sintel]{
    \includegraphics[width=0.46\linewidth]{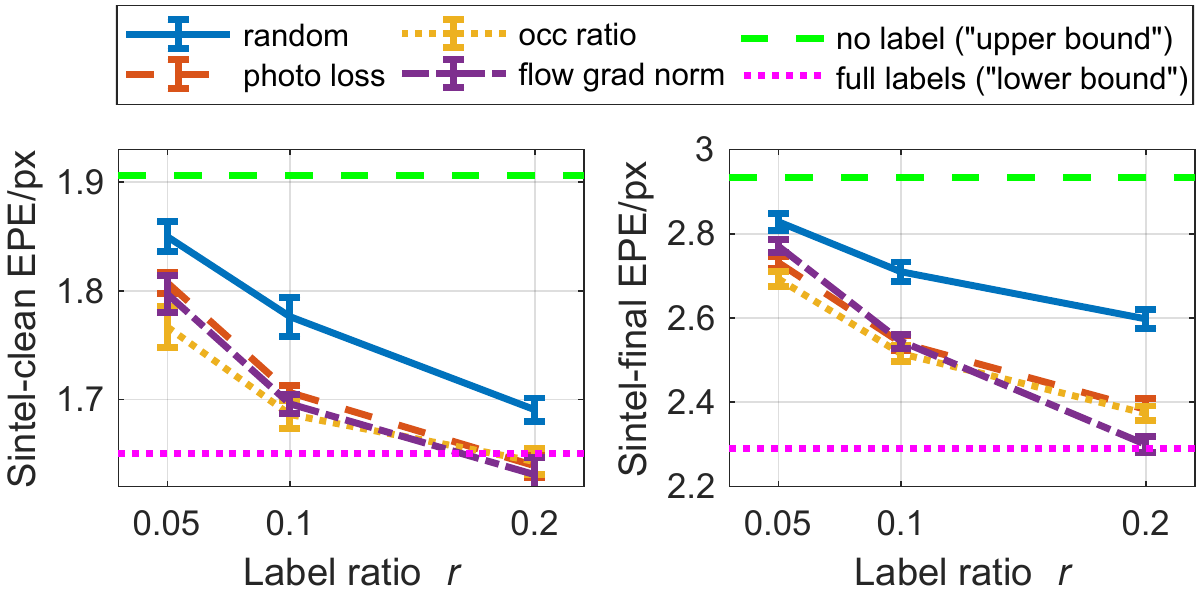} 
    \label{fig:sintel_active}
  }
  \subfigure[KITTI]{
    \includegraphics[width=0.46\linewidth]{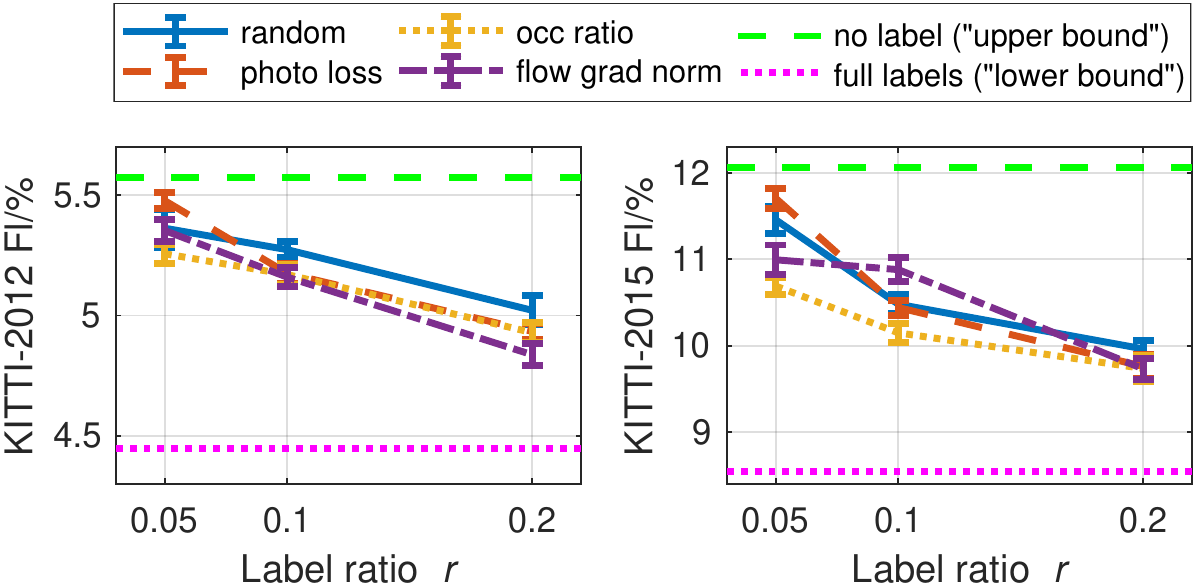} 
    \label{fig:kitti_active}
  }   
  \caption{Validation errors of different active learning algorithms compared with random sampling (baseline); pseudo error bars obtained by taking the standard deviations in the last 50 epoches}
  \label{fig:exp2}
\end{figure}

Our active learning results are shown in \cref{fig:exp2}. We compare the validation errors for our three active learning criteria against the baseline setting, in which the labeled samples are selected randomly. To better illustrate the scale of the differences, we add two horizontal lines to indicate totally unsupervised and supervised errors as the ``upper'' and ``lower'' bound, respectively.

The Sintel results (\cref{fig:sintel_active}) show that all our three active learning algorithms can improve the baseline errors by large margins. Notably, our active learning algorithms can achieve close to supervised performance with only 20\% labeling. This number is around 50\% without active learning.

The KITTI results (\cref{fig:kitti_active}) show slight improvements with active learning. Among our three algorithms, ``occ ratio'' works consistently better than random sampling, especially at a very small label ratio $r=0.05$. We discuss the reason why our active learning methods help less on KITTI at the end of this chapter.

Among our three active learning heuristics, ``occ ratio'' has the best performance overall and is therefore selected as our final criterion. Note that the occlusion ratio is computed via a forward-backward consistency check, so it captures not only real occlusions but also inconsistent flow estimates. 

\paragraph{Benchmark Testing} We also show results on the official benchmark test sets. Qualitative examples are also included in the appendix. As is shown in \cref{tab:sintel_test}, compared with the backbone ARFlow~\cite{liu2020learning} and two other top unsupervised estimators~\cite{jonschkowski2020matters,liu2019selflow}, our Sintel test EPEs improve significantly even when we utilize a very small fraction (5-20\%) of labels in training. This holds true for both clean and final passes, as well as occluded and non-occluded pixels. To indicate the scale of improvements, our semi-supervised results are even comparable to the supervised IRR-PWC~\cite{hur2019iterative}, which has a similar PWC-Net-based structure, even if we only use 20\% of the Sintel labels. We also include the state-of-the-art RAFT~\cite{hur2019iterative} results to get a sense of the overall picture.

In addition, \cref{tab:sintel_test} also shows that our active learning method works favorably against the baseline (``rand''). We found that our active learning method (``occ'') may overly sample the same scenes (\eg ``ambush''), so we also test an alternative (``occ-2x'') to balance the queried samples. Specifically, we select a double number of samples with top uncertainties and then randomly sample a half from them to query labels. This helps diversify our selected samples when the label ratio is very small. Our active learning methods perform comparably or better than the baseline, especially on the realistic final pass.

\setlength{\tabcolsep}{3pt}
\begin{table}[tbp]
\centering
\small
\caption{Sintel benchmark results (EPE/px). Metrics evaluated at `all' (all pixels), `noc' (non-occlusions), and `occ' (occlusions). The key metrics (used to sort on the official website) are underlined.
Parenthesis means evaluation data used in training. For all metrics, lower is better.}
\label{tab:sintel_test}
\begin{tabular}{ccl|cc|ccc|ccc}
\hline
\multicolumn{2}{c}{\multirow{3}{*}{Label ratio $r$}} &
  \multicolumn{1}{c|}{\multirow{3}{*}{Method}} &
  \multicolumn{2}{c|}{Train} &
  \multicolumn{6}{c}{Test} \\ 
\multicolumn{2}{c}{}                          & \multicolumn{1}{c|}{} & Clean & Final & \multicolumn{3}{c|}{Clean} & \multicolumn{3}{c}{Final} \\
\multicolumn{2}{c}{}                          & \multicolumn{1}{c|}{} & all   & all   & \underline{{all}}    & noc    & occ     & \underline{{all}}    & noc    & occ     \\ \hline
\multirow{3}{*}{\rotatebox[origin=c]{90}{unsup}}    & \multirow{3}{*}{$r=0$}    & SelFlow~\cite{liu2019selflow} & (2.88)  & (3.87)  & 6.56 & 2.67 & 38.30 & 6.57 & 3.12 & 34.72 \\
 &                      & UFlow~\cite{jonschkowski2020matters}   & (2.50)          & (3.39)        & 5.21  & 2.04  & 31.06  & 6.50  & 3.08  & 34.40  \\
 &                      & ARFlow~\cite{liu2020learning}  & (2.79)         & (3.73)        & \textbf{4.78}  & \textbf{1.91}  & \textbf{28.26}  & \textbf{5.89}  & \textbf{2.73}  & \textbf{31.60}  \\ \hline
\multirow{9}{*}{\rotatebox[origin=c]{90}{semi-sup}} & \multirow{3}{*}{$r=0.05$} & Ours(rand)  & (2.09) & 
(2.99) & 4.04 & 1.52 & 24.65 & 5.49 & 2.62 & 28.86 \\
 &                      & Ours(occ)    & (1.95)        & (2.38)       & 4.11  & 1.63  & \textbf{24.39}  & \textbf{5.28}  & \textbf{2.49}  & \textbf{28.03}   \\
 &                      & Ours(occ-2x)  & (1.94)        & (2.55)       & \textbf{3.96}  & \textbf{1.45}  & 24.42  & 5.35  & 2.50  & 28.58  \\ \cline{3-11} 
 & \multirow{3}{*}{$r=0.1$} & Ours(rand)  & (2.36)        & (3.18)        & \textbf{3.91}  & \textbf{1.47}  & \textbf{23.82}  & 5.21  & 2.46  & 27.66  \\
 &                      & Ours(occ)     & (1.64)        & (1.98)       & 4.28  & 1.68  & 25.49  & 5.31  & \textbf{2.44}  & 28.68  \\
 &                      & Ours(occ-2x)  & (1.75)        & (2.30)       & 4.06  & 1.63  & 23.94  & \textbf{5.09}  & 2.49   & \textbf{26.31}  \\ \cline{3-11} 
 & \multirow{3}{*}{$r=0.2$} & Ours(rand)  & (2.17)        & (2.93)       & 3.89  & 1.56  & \textbf{22.86}  & 5.20  & 2.50  & 27.19   \\
 &                      & Ours(occ)     & (1.35)         & (1.63)       & 4.36  & 1.86  & 24.76   & 5.09  & 2.45  & 26.69  \\
 &                      & Ours(occ-2x)  & (1.57)        & (2.05)        & \textbf{3.79}   & \textbf{1.44}  & 23.02  & \textbf{4.62}  & \textbf{2.07}   & \textbf{25.38}  \\ \hline
\multirow{3}{*}{\rotatebox[origin=c]{90}{sup}}      & \multirow{3}{*}{$r=1$}    & PWC-Net~\cite{sun2018pwc} & (2.02)  & (2.08)  & 4.39 & 1.72 & 26.17 & 5.04 & 2.45 & 26.22 \\
 &                      & IRR-PWC~\cite{hur2019iterative} & (1.92)         & (2.51)        & 3.84  & 1.47  & 23.22   & 4.58  & 2.15  & 24.36  \\
 &                      & RAFT~\cite{teed2020raft}    & (0.77)         & (1.27)        & \textbf{1.61}  & \textbf{0.62}  & \textbf{9.65}   & \textbf{2.86}  & \textbf{1.41}  & \textbf{14.68}   \\ \hline
\end{tabular}
\end{table}
\setlength{\tabcolsep}{1.4pt}

\Cref{tab:kitti_test} shows our benchmark testing results on KITTI. Consistent with our findings on Sintel, our semi-supervised methods are significantly better than the compared unsupervised state-of-the-art methods, and close to the supervised IRR-PWC~\cite{hur2019iterative}, even if we only use a very small fraction (5-20\%) of labels. In addition, our active learning method also works consistently better than the baseline for all tested label ratios, especially on the harder KITTI-2015 set. 

\setlength{\tabcolsep}{2.4pt}
\begin{table}[tbp]
\centering
\small
\caption{KITTI benchmark results (EPE/px and Fl/\%). Metrics evaluated at `all' (all pixels, default for EPE), `noc' (non-occlusions), `bg' (background), and `fg' (foreground). Key metrics (used to sort on the official website) are underlined.
`()' means evaluation data used in training. `-' means unavailable. For all metrics, lower is better.}
\label{tab:kitti_test}
\begin{tabular}{ccl|cc|cc|cccc}
\hline
\multicolumn{2}{c}{\multirow{3}{*}{Label ratio $r$}} &
  \multicolumn{1}{c|}{\multirow{3}{*}{Method}} &
  \multicolumn{2}{c|}{Train} &
  \multicolumn{6}{c}{Test} \\ 
\multicolumn{2}{c}{}                          & \multicolumn{1}{c|}{} & 2012 & 2015 & \multicolumn{2}{c|}{2012} & \multicolumn{4}{c}{2015} \\
\multicolumn{2}{c}{}                          & \multicolumn{1}{c|}{} & EPE   & EPE   & \underline{{Fl-noc}}    & EPE    & \underline{{Fl-all}}     & Fl-noc    & Fl-bg    & Fl-fg     \\ \hline
\multirow{3}{*}{\rotatebox[origin=c]{90}{unsup}}    & \multirow{3}{*}{$r=0$}    & SelFlow~\cite{liu2019selflow} & (1.69)  & (4.84)  & 4.31 & 2.2 & 14.19 & 9.65 & 12.68 & 21.74 \\
 &                      & UFlow~\cite{jonschkowski2020matters}   & (1.68)          & (2.71)        & 4.26  & 1.9  & 11.13  & 8.41  & 9.78  & 17.87  \\
 &                      & ARFlow~\cite{liu2020learning}  & (1.44)         & (2.85)        & -  & 1.8  & 11.80  & -  & -  & -  \\ \hline
\multirow{6}{*}{\rotatebox[origin=c]{90}{semi-sup}} & \multirow{2}{*}{$r=0.05$} & Ours(rand)  & (1.25) & 
(2.61) & 3.90 & 1.6 & 9.77 & 6.99 & 8.33 & 17.02 \\
 &                      & Ours(occ)    & (1.22)        & (2.29)       & 3.90  & \textbf{1.5}  & \textbf{9.65}  & \textbf{6.94}  & \textbf{8.20}  & \textbf{16.91}   \\ \cline{3-11} 
 & \multirow{2}{*}{$r=0.1$} & Ours(rand)  & (1.21)        & (2.56)        & 3.75  & 1.5  & 9.51  & 6.69  & 8.01  & 17.01  \\
 &                      & Ours(occ)     & (1.21)        & (1.98)       & \textbf{3.74}  & 1.5  & \textbf{8.96}  & \textbf{6.28}  & \textbf{7.74}  & \textbf{15.04}  \\ \cline{3-11} 
 & \multirow{2}{*}{$r=0.2$} & Ours(rand)  & (1.16)        & (2.10)       & 3.50  & 1.5  & 8.38  & \textbf{5.68}  & 7.37  & \textbf{13.44}   \\
 &                      & Ours(occ)     & (1.13)         & (1.73)       & \textbf{3.49}  & 1.5  & \textbf{8.30}   & 5.69  & \textbf{7.25}  & 13.53  \\ \hline
\multirow{3}{*}{\rotatebox[origin=c]{90}{sup}}      & \multirow{3}{*}{$r=1$}    & PWC-Net~\cite{sun2018pwc} & (1.45)  & (2.16)  & 4.22 & 1.7 & 9.60 & 6.12 & 9.66 & 9.31 \\
 &                      & IRR-PWC~\cite{hur2019iterative} & -         & (1.63)        & 3.21  & 1.6  & 7.65   & 4.86  & 7.68  & 7.52  \\
 &                      & RAFT~\cite{teed2020raft}    & -         & (0.63)        & -  & -  & \textbf{5.10}   & \textbf{3.07}  & \textbf{4.74}  & \textbf{6.87}   \\ \hline
\end{tabular}
\end{table}
\setlength{\tabcolsep}{1.4pt}


\paragraph{Ablation Study on Settings of Stage 2}

We try different active learning schedules in Stage 2 and show our current setting works the best. We report the Sintel final EPE for different Stage 2 settings with label ratio $r=0.1$. In \cref{tab:sintel_ablation}, the first row is our current Stage 2 setting, \ie semi-supervised training on the partial labeled train set. The second row refers to supervised training only on the labeled part of train set, without the unsupervised samples. The third row considers also including the unlabeled raw data (used in Stage 1) in the Stage 2 semi-supervised training. We can see that our current setting works significantly better than the two alternatives. The second setting works poorly due to overfitting on the very small labeled set, which means that the unlabeled part of the train split helps prevent overfitting. The third setting also fails due to the excessive amount of unlabeled data used in Stage 2, which overwhelms the small portion of supervised signal from queried labels.

\setlength{\tabcolsep}{4pt}
\begin{table}[tbp]
\caption{Ablation study: different Stage 2 settings. Sintel final validation EPE, label ratio $r=0.1$. Standard deviations from the last 50 epoches. * denotes current setting.}
\label{tab:sintel_ablation}
\centering
\small
\begin{tabular}{c|cccc}
\hline
\multirow{2}{*}{Data split {[}Loss{]}} & \multicolumn{4}{c}{Method}                       \\
                                       & random & photo loss & occ ratio & flow grad norm \\ \hline
train {[}semi-sup{]*}     & \textbf{2.71($\pm$0.02)} & \textbf{2.54($\pm$0.02)} & \textbf{2.52($\pm$0.02)} & \textbf{2.54($\pm$0.02)} \\
train {[}sup{]}          & 2.82($\pm$0.01) & 2.82($\pm$0.01) & 2.59($\pm$0.01) & 2.77($\pm$0.01) \\
raw+train {[}semi-sup{]} & 3.13($\pm$0.04)  & 3.09($\pm$0.05)  & 3.15($\pm$0.06)  & 3.07($\pm$0.05) \\ \hline
\end{tabular}
\end{table}
\setlength{\tabcolsep}{1.4pt}


\paragraph{Model Analysis and Visualization}

\begin{figure}[tbp]
\centering
\subfigure[Selected samples in Sintel]{
    \includegraphics[width=0.37\linewidth]{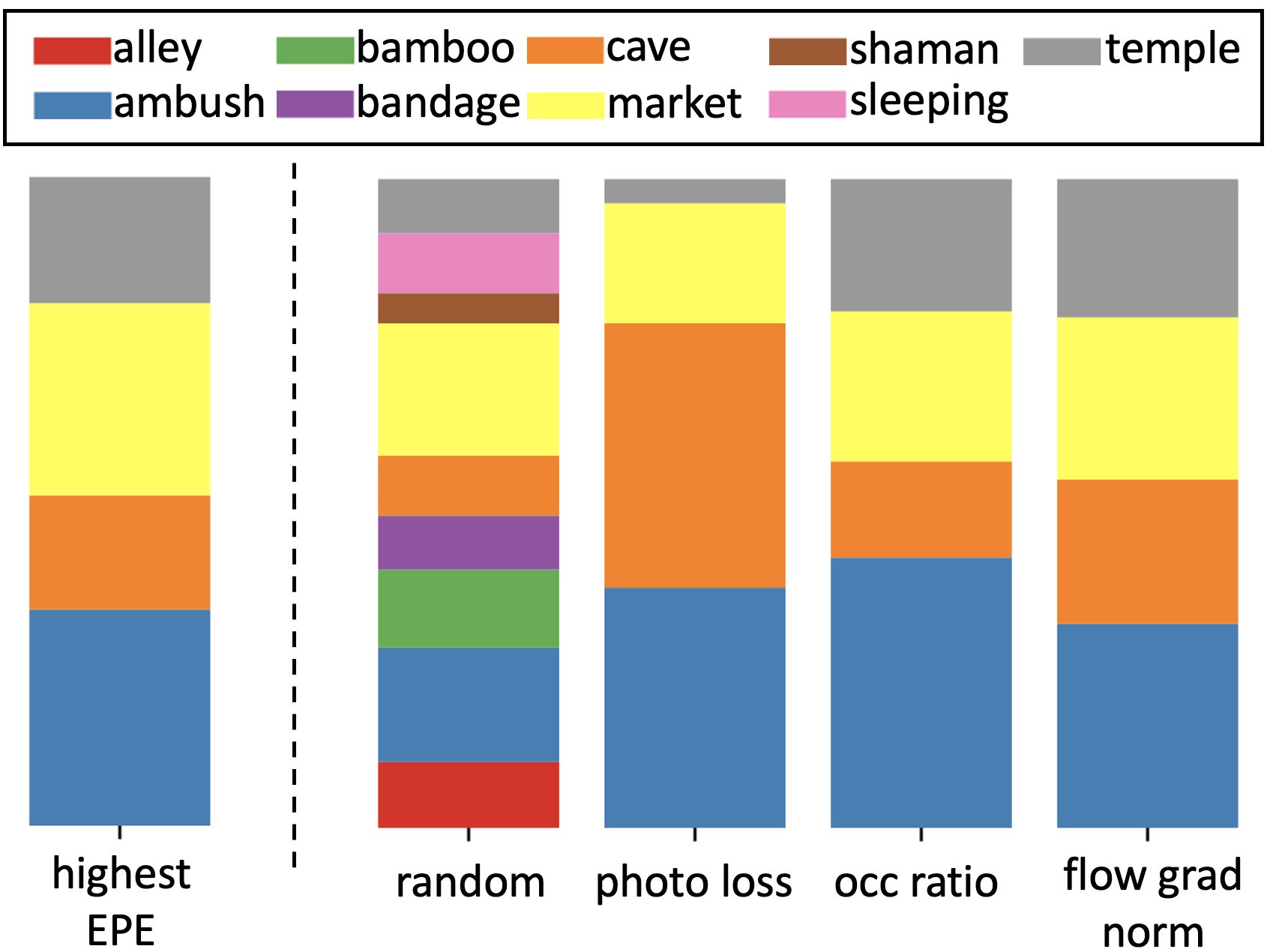}
    \label{fig:active_learning_analysis_a}
    }
\subfigure[Sintel]{
    \includegraphics[width=0.31\linewidth]{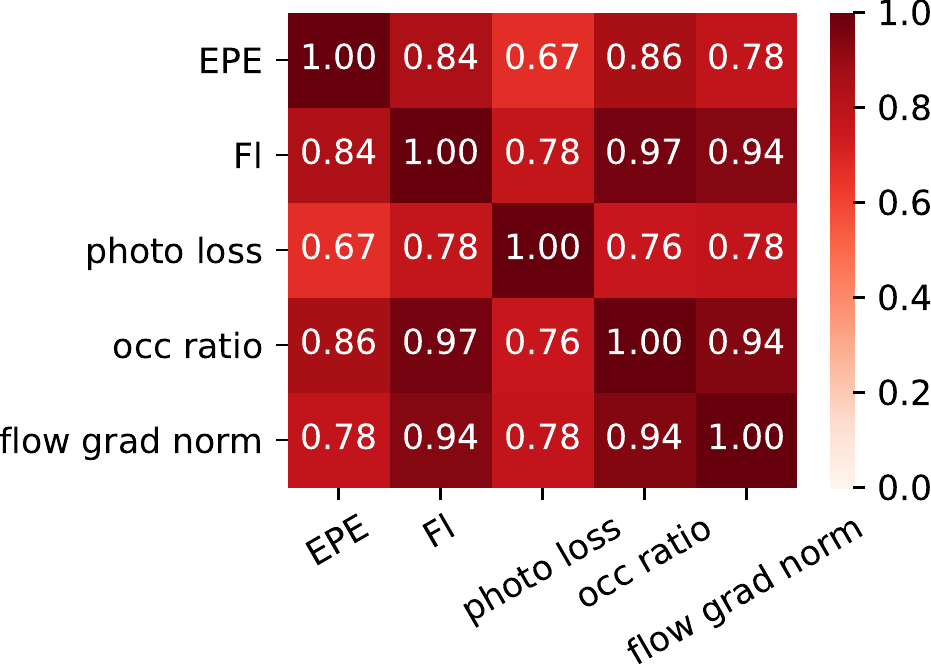}
    \label{fig:active_learning_analysis_b}
    }
\subfigure[KITTI]{    
    \includegraphics[width=0.22\linewidth]{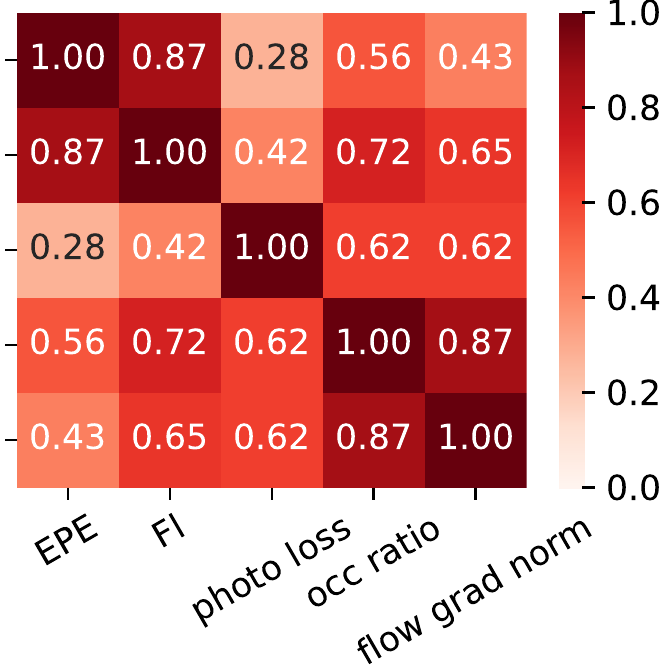}
    \label{fig:active_learning_analysis_c}
    }
\caption{(a) Sintel samples selected by different methods ($r=0.2$), grouped by scenes; and correlation matrices with sample errors for Sintel (b) and KITTI (c).}
\label{fig:active_learning_analysis}
\end{figure}

\Cref{fig:active_learning_analysis_a} shows which Sintel samples are selected by different active learning methods. As shown in the left-most column, the pre-trained model after Stage 1 generally has high EPEs (top 20\% shown in the figure) on four scenes, namely ``ambush'', ``cave'', ``market'', and ``temple''. The random baseline tends to select a bit of every scene, whereas all our three active learning algorithms query scenes with high EPEs for labels. This confirms that our active learning criteria capture samples that are especially challenging to the current model, which explains the success of active learning.

We also analyze the relationships between our criteria and model errors through correlation matrices visualized by heat maps in \cref{fig:active_learning_analysis_b,fig:active_learning_analysis_c}. We can see that the sample errors in Sintel generally have high correlations with all three score values, whereas in KITTI the correlations are much smaller. Also, the ``occ ratio'' score generally has the highest correlation with sample errors among the three proposed methods. All these observations are consistent with our active learning validation results. Thus, we posit that the correlation between uncertainty values and sample errors can be a good indicator in designing effective active learning criteria.

\paragraph{Discussion on Factors That May Influence Active Learning} 

\begin{figure}[t]
    \centering
    \includegraphics[width=\linewidth]{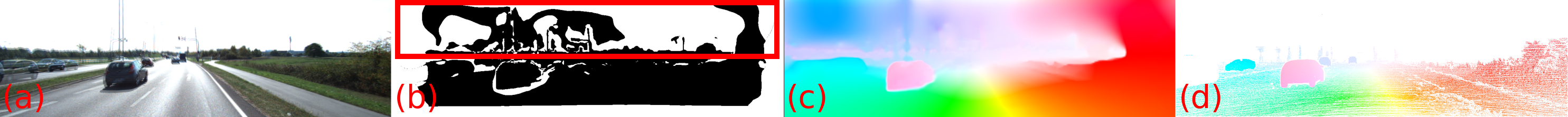}
    \caption{An example for the information mismatch problem. Data from KITTI-2015, frame 79 (Fl=19.35\%), with the third largest ``occ ratio'' score: (a) superposed input images; (b) estimated occlusion map; (c) flow prediction; (d) flow ground truth.}
    \label{fig:active_mismatch_example_79}
\end{figure}

\begin{itemize}
    \item \textbf{Pattern Homogeneity}: Based on our validation results in \cref{fig:exp2}, active learning seems more effective on Sintel than on KITTI. This may be because KITTI samples are relatively more homogeneous in terms of motion patterns. Unlike the Sintel movie sequences, which contain arbitrary motions of various scales, driving scenes in KITTI exhibit a clear \emph{looming motion} caused by the dominant forward motion of the vehicle that carries the camera. Specifically, Sintel has extremely hard scenes like ``ambush'' as well as extremely easy scenes like ``sleeping''. This large variation of difficulty makes it possible to select outstandingly helpful samples and labels. In contrast, since KITTI motions are more patterned and homogeneous, any selection tends to make little difference with respect to random sampling.

    \item  \textbf{Label Region Mismatch}: KITTI only has sparse labels, \ie only a part of the image pixels have labels. This is crucial because our active learning criteria are computed over the whole frame, so there is a mismatch between the support of our criteria and the KITTI labels. Specifically, the sparse labels may not cover the problematic regions found by our criteria. One example is shown in \cref{fig:active_mismatch_example_79}. The sky region has bad predictions due to lack of texture, and the ``occ ratio'' method captures the inconsistent flow there by highlighting the sky region. However, the ground-truth labels do not cover the sky region, so having this sample labeled does not help much in training.
    
\end{itemize}

\section{Conclusion} \label{sec:conclusion}

In this paper, we first analyzed the trade-off between model performance and label ratio using a simple yet effective semi-supervised optical flow network and found that the unsupervised performance can be significantly improved even with a small fraction of labels. We then explored active learning as a way to further improve the performance and reduce annotation costs. Our active learning method works consistently better than baseline on Sintel and KITTI datasets.

For potential future work, it may be interesting to explore how to deal with sparse labels in the active learning framework or how to query labels by region rather than full frame.

\paragraph{Acknowledgments.}This material is based upon work supported by the National Science Foundation under Grant No. 1909821 and by the Intelligence Advanced Research Projects Agency under contract number 2021-21040700001.


\clearpage
%
%
\bibliographystyle{splncs04}
\bibliography{main}

\appendix
\begin{center}
    \Large \textbf{Appendix}
\end{center}
\paragraph{Table of Contents}

\begin{itemize}
    \item[\ref{app:method_details}] \textbf{Method Details} \dotfill \pageref{app:method_details}
    
    \begin{itemize}
        \item[\ref{app:semi-supervised_training}] Semi-supervised training \dotfill \pageref{app:semi-supervised_training}
        \item[\ref{app:network_choice}] Network choice: why ARFlow? why not RAFT? \dotfill \pageref{app:network_choice}
    \end{itemize}
    
    \item[\ref{app:experiment_details}] \textbf{Experiment Details} \dotfill \pageref{app:experiment_details}
    
    \begin{itemize}
        \item[\ref{app:summary_of_available_datasets}] Summary of Available Datasets \dotfill \pageref{app:summary_of_available_datasets}
        \item[\ref{app:data_augmentation_parameters}] Data Augmentation Parameters \dotfill \pageref{app:data_augmentation_parameters}
        \item[\ref{app:training_schedule_design}] Training Schedule Design \dotfill \pageref{app:training_schedule_design}
        \item[\ref{app:note_on_sintel}] A Special Note on Sintel Label Queries in Pairs \dotfill \pageref{app:note_on_sintel}
    \end{itemize}    
    
    \item[\ref{app:more_data_and_results}] \textbf{More Data and Results} \dotfill \pageref{app:more_data_and_results}
    
    \begin{itemize}
        \item[\ref{app:raw_validation_data}] Raw Validation Data \dotfill \pageref{app:raw_validation_data}
        \item[\ref{app:benchmark_qualitative_results}] Benchmark Qualitative Results \dotfill \pageref{app:benchmark_qualitative_results}
        \item[\ref{app:analysis_on_more_uncertainty_scores}] Analysis on More Uncertainty Scores \dotfill \pageref{app:analysis_on_more_uncertainty_scores}
    \end{itemize}

\end{itemize}

\section{Methodology Details}\label{app:method_details}



\subsection{Semi-supervised Training}\label{app:semi-supervised_training}

In this work, we explore the spectrum between totally supervised (100\% labeled) and totally unsupervised (0\% labeled) training. The question is: what is the intermediate state of semi-supervised learning if we have exactly a fraction $r$ of training samples labeled ($0<r<1$)? Intuitively, the performance should be monotonically increasing when we increase the label ratio $r$.

Specifically, the most ideal setting requires us to find a semi-supervised learning scheme that 
\begin{itemize}
    \item trains using the information of all labeled and unlabeled samples at the same time, and
    \item is continuous at $r=0$ (unsupervised) and $r=1$ (supervised), \ie, if we set $r=0$, it should be equivalent as the current unsupervised learning pipeline, and if we set $r=1$, it should be equivalent as the fully supervised setting.
\end{itemize}

We want to define our semi-supervised setting as a smooth transition between the supervised and unsupervised settings. The naive solution is to train a fixed neural network architecture with a semi-supervised loss that works differently for the labeled and unlabeled samples. For example,

\begin{equation}
\ell_{\text{semi}}(\bm x) = \left\{\begin{array}{ll}
\ell_{\text{unsup}}(\bm x),   & \text{if $\bm x$ is unlabeled,} \\
\alpha\ell_{\text{sup}}(\bm x),   & \text{otherwise,}
\end{array}\right.
\end{equation}
where $\alpha>0$ is the coefficient to balance the two different losses. We then have the final loss term
\[
\cL_{\text{semi}} = \sum_{\bm x\in \cD} \ell_{\text{semi}}(\bm x) = \sum_{\bm x\in \cD^{u}} \ell_{\text{unsup}}(\bm x) + \alpha \sum_{\bm x\in \cD^{l}} \ell_{\text{sup}}(\bm x),
\]
where $\cD^u$ and $\cD^l$ are the unlabeled and labeled sample set, and $\cD=\cD^u\cup\cD^l$. Under this setting, the label ratio is $r=|\cD^l|/(|\cD^u|+|\cD^l|)$, and changing $r$ from 0 to 1 will change the setting smoothly from unsupervised to supervised.

However, one concern with this setting is that we need to fix the same dataset $\cD$ for all experiments with varying $0\leq r\leq 1$, but the datasets used in current state-of-the-art supervised and unsupervised methods are usually different. For example, to get the best results on the Sintel dataset~\cite{sintel}, unsupervised methods first train on the Sintel raw movie dataset and then fine-tune on Sintel. However, the latest supervised methods usually first train on the FlyingChairs~\cite{dosovitskiy2015flownet} and FlyingThings3D~\cite{flyingthings3D} datasets before training on the small Sintel set. This difference in dataset is important to notice because it is one of the advantages of unsupervised learning that it can use much more data (probably from the same data distribution as the test data) than supervised training.

In light of the problem mentioned above, we decide to use the unsupervised datasets as our data in the semi-supervised training. There are mainly two reasons. First of all, the label ratio can be very low in daily practice, so defining our setting closer to the unsupervised setting may be more practical. Second, the unsupervised training is harder to converge than the supervised training because of the lack of supervisory signals, so using a framework that is closer to the unsupervised training may be better for convergence in both scenarios. 

Another concern is in the training schedule. In the setting defined above, we only have one stage of training that use all labeled and unlabeled samples in the same stage. Another option is to split to two stages, one unsupervised stage using all unlabeled samples (similar as a pre-training stage) and a supervised stage using the labeled samples. We discuss the pros and cons of both schedules in \cref{app:training_schedule_design}.

\subsection{Network Choice: Why ARFlow? Why Not RAFT?}\label{app:network_choice}

We found that RAFT is not appropriate to be tested at this stage because it has been mostly proven to work in the supervised setting, but our semi-supervised flow is actually much closer to the unsupervised setting in the following two ways.
\begin{itemize}
    \item Our label ratio is very low (5-10\%), which is almost unsupervised. The supervision signal is extremely sparse.
    \item Our first training stage is unsupervised, so the model is initialized in an unsupervised way.
\end{itemize}
Therefore, a reliable unsupervised base model is preferred in our setting. This is why we choose ARFlow~\cite{liu2020learning} (unsupervised SOTA) instead of RAFT~\cite{teed2020raft} (supervised SOTA).

Admittedly, there is recent work~\cite{stone2021smurf} on unsupervised versions of RAFT. However, this work is based on multi-frame inputs, and it also adds too much complexity (such as self-supervision) into the model, so we do not think it is the right time to move towards RAFT now. However, we do agree that it is worth trying in the future once a simple and reliable unsupervised appraoch for RAFT is available.

\section{Experiment Details}\label{app:experiment_details}

\subsection{Summary of Available Datasets}\label{app:summary_of_available_datasets}

\paragraph{Official Datasets} We train and evaluate our method on two large synthetic datasets, FlyingChairs~\cite{dosovitskiy2015flownet} and FlyingThings3D~\cite{flyingthings3D}, as well as two more realistic datasets, Sintel~\cite{sintel} and KITTI~\cite{kitti12,kitti15}.

FlyingChairs~\cite{dosovitskiy2015flownet} and FlyingThings3D~\cite{flyingthings3D} consist of image pairs generated by moving chairs or everyday objects across the background images along randomized 3D trajectories. These two datasets are large but unrealistic, so they are usually only used to pre-train supervised networks.

Sintel~\cite{sintel} is a challenging benchmark dataset obtained from a computer-animated movie. This dataset is closer to real-life scenes as it contains fast motions, large occlusions, and many realistic artifacts like illumination change and fog or blur. It provides both clean and final passes with corresponding dense optical flow labels. Apart from that, the unlabeled raw movie frames have also been used in many recent unsupervised work~\cite{liu2019selflow,liu2020learning}.

KITTI dataset was first released in 2012~\cite{kitti12} and extended in 2015\cite{kitti15}. The dataset contains frame pairs of road scenes from a camera mounted on a car. Sparse optical flow labels are provided using 3D laser scanner and egomotion information. KITTI raw frames with no labels are also available and used in unsupervised training~\cite{ranjan2019competitive,liu2020learning,yin2018geonet}. 

\cref{tab:available_data} summarizes the dataset information. We have excluded the labeled samples from the raw Sintel and KITTI dataset, so all splits in the table are disjoint.

\begin{table}[ht]
\centering
\small
\caption{Available official datasets for optical flow estimation.}\label{tab:available_data}
\begin{tabular}{c|c|cc}
\hline
Dataset                  & Split           & \# of samples & Labeled? \\ \hline
\multirow{2}{*}{FlyingChairs}                    & train           & 22,232         & \cmark         \\
                   & val           & 640         & \cmark         \\ \hline
\multirow{2}{*}{FlyingThings3D} & train           & 19,621         & \cmark         \\
                          & val             & 3,823          & \cmark         \\ \hline
\multirow{3}{*}{Sintel}   & raw             & 12,466         & \xmark         \\
                          & clean     & 1,041          & \cmark         \\
                          & final     & 1,041          & \cmark         \\ \hline
\multirow{3}{*}{KITTI}    & raw             & 27,858         & \xmark         \\
                          & 2012      & 194           & \cmark         \\
                          & 2015      & 200           & \cmark         \\ \hline
\end{tabular}
\end{table}

\paragraph{Our Data Splits} We cannot test our model using the official KITTI and Sintel test set for all the experiments since the website restricts the number of submissions. We test on the official test set only for the major final models in our paper. Thus, we need to split our own validation set from Sintel and KITTI.

Our own train/val split is shown in Tab \ref{tab:our_split}. For Sintel, as suggested in the official implementation of ARFlow~\cite{liu2020learning}, we split the following folders of both clean and final passes as our \emph{train} split of Sintel: \text{alley\_1}, \text{ambush\_4}, \text{ambush\_6}, \text{ambush\_7}, \text{bamboo\_2}, \text{bandage\_2}, \text{cave\_2}, \text{market\_2}, \text{market\_5}, \text{shaman\_2}, \text{sleeping\_2}, and \text{temple\_3}. For KITTI, we take the first 150 samples in each of the 2015 set and 2012 set as our \emph{train} split and the rest as our \emph{val} split. 

\begin{table*}[t]
\centering
\caption{Our train/val split of Sintel and KITTI}\label{tab:our_split}
\begin{tabular}{c|c|cc}
\hline
Dataset   & Our train split                       &\# train samples  & \# val samples       \\ \hline
\multirow{4}{*}{Sintel clean+final} & alley\_1, ambush\_4, ambush\_6,       & \multirow{4}{*}{1082} & \multirow{4}{*}{1000} \\
                                           & ambush\_7, bamboo\_2, bandage\_2,     &                   &                       \\
                                           & cave\_2, market\_2, market\_5,        &                   &                       \\
                                           & shaman\_2, sleeping\_2, and temple\_3 &                   &                       \\ \hline
KITTI 2015+2012           & first 150 samples for each            & 300               & 94                    \\ \hline
\end{tabular}
\end{table*}

\subsection{Data Augmentation Parameters}\label{app:data_augmentation_parameters}

\begin{table*}[t]
\centering
\small
\caption{Data augmentation parameters}\label{tab:data_augmentation}
\begin{tabular}{c|c|c|c|c}
\hline
                           & FlyingChairs & {FlyingThings3D} & {Sintel} & {KITTI} \\ \hline
Cropping            & $384\times448$        &  $384\times768$         &  $384\times768$               &   $320\times960$              \\ \hline
\multirow{2}{*}{Rescaling} & \multirow{2}{*}{\xmark}       & \multirow{2}{*}{\xmark} & \multicolumn{1}{c|}{scale $\in[2^{-0.2}, 2^{0.6}]$}                   & \multirow{2}{*}{\xmark} \\
                           &  & & \multicolumn{1}{c|}{with prob. 0.8} &    \\ \hline
Horizontal flip        & with prob. 0.5        &   with prob. 0.5        &   with prob. 0.5                &       with prob. 0.5            \\ \hline
\multirow{6}{*}{Appearance} & brightness = 0.5    & brightness = 0.5      &  brightness = 0.4 &  brightness = 0.3  \\ 
                                            & contrast = 0.5 &  contrast = 0.5     &  contrast = 0.4 &  contrast = 0.3  \\
                                            & saturation = 0.5    &   saturation = 0.5    &  saturation = 0.4 &  saturation = 0.3   \\
                                            & hue = 0    &  hue = 0     & hue = 0.16 & hue = 0.1  \\     
                                            & gamma = \texttt{True}  &  gamma = \texttt{True}     & gamma = \texttt{True} & gamma = \texttt{False } \\
                                            & gblur = \texttt{True} &  gblur = \texttt{True}     &  gblur = \texttt{True}   &  gblur = \texttt{True}  \\ \hline     
\end{tabular}
\end{table*}

The data augmentation parameters in our experiments are summarized in \cref{tab:data_augmentation}. Our data augmentation implementations are borrowed from the official code base of ARFlow~\cite{liu2020learning} and RAFT~\cite{teed2020raft}. We use ColorJitter from the \text{torchvision.transforms} package to implement the appearance transformations. In addition, ``gamma'' means raising the normalized image color value (between 0 and 1) to a power sampled between 0.7 and 1.5 uniformly, and ``gblur'' means applying gaussian blur with radius 3 with probability 0.5.

\subsection{Training Schedule Design}\label{app:training_schedule_design}

Three options of the training pipelines are listed below. We now explain and discuss them one by one.

\begin{itemize}
\item[A.] train on \emph{all} data (semi-sup)
\item[B.] train on \emph{all} data (unsup) $\rightarrow$ query partial labels from \emph{all} data $\rightarrow$ train on \emph{all} data (semi-sup)
\item[C.] train on \emph{non-candidate} set (unsup) $\rightarrow$ query partial labels from \emph{candidate} set $\rightarrow$ train on \emph{candidate} set (semi-sup)
\end{itemize}

A one-stage training schedule (option A) can be used if our goal is only to visualize the change of performance when we gradually increase the label ratio from 0 to 1. We can simply train on the full dataset with partial labels using the semi-supervised loss in one stage. Thus, we use this setting in our first experiment to draw the label ratio-validation error curves. We randomly shuffle and mix labeled and unlabeled samples in mini-batches to stabilize our training. However, this assumes that the partial labels have to be assigned \emph{before} training independent of the model and thus may be naive compared with the other two options, which use active learning. 

Now, we want to explore a semi-supervised training pipeline where the labels are assigned \emph{during} the training. This has to be a pipeline of at least two stages because we need to query labels at some point in the process. Specifically, as shown in option B, we first have a totally unlabeled dataset, so we train our first model using the unsupervised loss. Then, based on the current trained model, we pick a part of the dataset that can help the current model most to query labels. Subsequently, we continue training using the semi-supervised loss. This reflects a workflow that can be applied in real practice so that the researchers only need to pay for the partial labels that can help the most. Note that we can easily change the pipeline to query labels multiple times by stacking more stages in the end.

One problem for option B is that it assumes every sample can be labeled. However, in real life, it is possible that only a subset of the original dataset can be labeled. For instance, labeling the ground-truth flow of an autonomous driving dataset (like KITTI) requires lidar sensors deployed when the videos are collected. If a raw video does not have the corresponding lidar information, it cannot be labeled but can still be used in the unsupervised part of training. Therefore, a more general setting is to define a candidate set to indicate those samples that can be labeled. Note that we can always split the full dataset manually to a candidate and a non-candidate splits even if every sample is eligible to get the label, which actually brings benefits in generalization as we will discuss next.

After splitting the dataset to a candidate and a non-candidate set, we can define the pipeline as in option C above. We first do unsupervised training on the non-candidate set and then use the current model to select samples out of the candidate set to get labels. This is beneficial because the model has not seen the candidate set in its first stage of unsupervised training. This can help add generalization ability because when we select samples to label, we are actually validating the current model on the new unseen candidate set. The selected samples are thus the ones that can help the current model generalize the most. This is why we stick to option C as our experiment settings in all the experiments on KITTI and Sintel.

\paragraph{Experiments on semi-supervised training (drawing the label ratio-validation error curves)} Since our interest in this experiment is to see how the error changes when we assign different ratios of labels, we first use the simplest training schedule (option A) on two toy datasets, FlyingChairs and FlyingThings3D, to plot the whole figure. The experiment settings are as follows.

\begin{itemize}
\item FlyingChairs: train on the \emph{train} split (semi-sup), evaluate on the \emph{val} split
\item FlyingThings3D: train on the \emph{train} split (semi-sup), evaluate on the \emph{val} split
\end{itemize}

Subsequently, we also would like to see the curve on two regular datasets, Sintel and KITTI, but since a large part of the data (raw dataset) is not labeled, we have to pre-train on those data in an unsupervised manner. This fits into the reason for specifying a candidate set, where only part of the data we have in hand are eligible to query labels. Moreover, to better fit the state-of-the-art unsupervised training schedule, we adopt option C as our training schedule. For Sintel and KITTI, we assign our \emph{train} split as the candidate set, and the large unlabeled data (\emph{raw} sets) as the non-candidate set\footnote{We are not using the KITTI multi-view extension set for simplicity.}, yielding the following training schedules.

\begin{itemize}
\item Sintel: train on \emph{raw} Sintel videos (unsup) $\rightarrow$ \emph{randomly} select and assign labels for our \emph{train} split $\rightarrow$ train on our \emph{train} split (semi-sup) $\rightarrow$ evaluate on our \emph{val} split.
\item KITTI: train on \emph{raw} KITTI videos (unsup) $\rightarrow$ \emph{randomly} select and assign labels from our \emph{train} split $\rightarrow$ train on our \emph{train} split (semi-sup) $\rightarrow$ evaluate on our \emph{val} split.
\end{itemize}

\paragraph{Experiments on our active learning algorithms} We consider many heuristics as the algorithms to select samples to label. We use Sintel and KITTI datasets and apply the same training schedule as in the previous experiments. The only difference is that we now use our algorithms to select samples to label instead of random selection.

\begin{itemize}
\item Sintel: train on \emph{raw} Sintel videos (unsup) $\rightarrow$ apply our \emph{active learning algorithms} to select and assign labels for our \emph{train} split $\rightarrow$ train on our \emph{train} split (semi-sup) $\rightarrow$ evaluate on our \emph{val} split.
\item KITTI: train on \emph{raw} KITTI videos (unsup) $\rightarrow$ apply our \emph{active learning algorithms} to select and assign labels from our \emph{train} split $\rightarrow$ train on our \emph{train} split (semi-sup) $\rightarrow$ evaluate on our \emph{val} split.
\end{itemize}

\subsection{A Special Note on Sintel Label Queries in Pairs}\label{app:note_on_sintel}

When we run experiments on Sintel, the same set of labels are provided for both clean and final pass input frames, since the final pass is simply another rendering of the same content with more realistic artifacts like motion blur. In other words, we always ensure that the corresponding clean and final samples are either both labeled or both unlabeled. The reasons are as follows.

\begin{figure}[t] 
    \centering
    \subfigure[Sampling in pairs]{
        \includegraphics[width=0.4\linewidth]{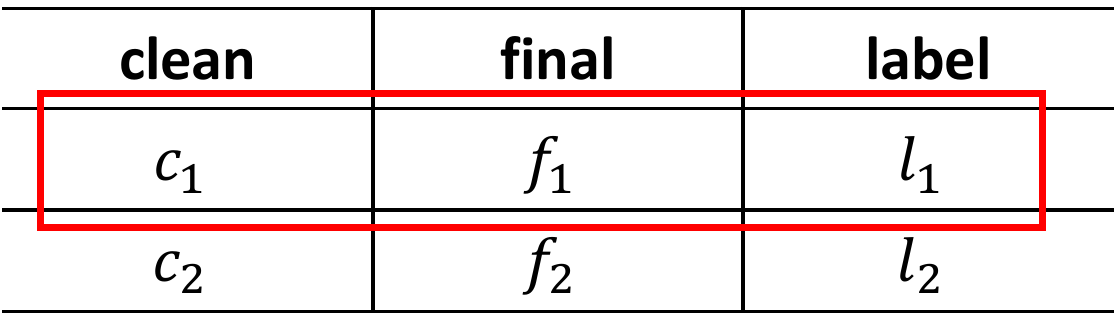} 
        \label{fig:sintel_pair1}
    }
    \subfigure[Sampling separately]{
        \includegraphics[width=0.4\linewidth]{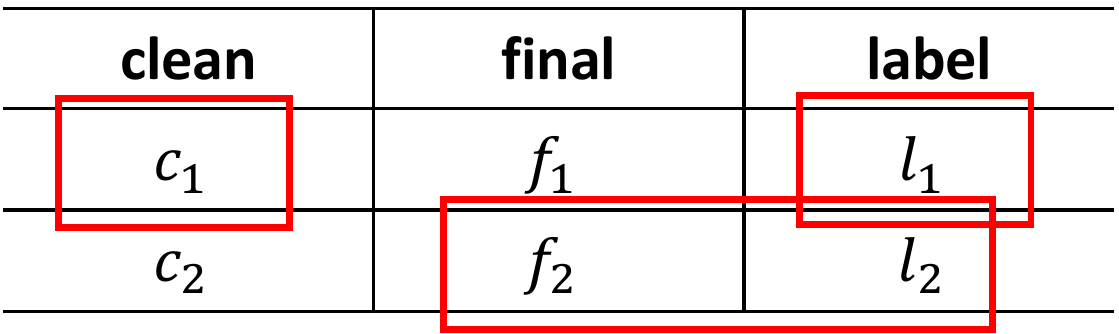} 
        \label{fig:sintel_pair2}
    }
    \caption{Examples of two different sampling methods on Sintel}
    \label{fig:sintel_pair}
\end{figure}

In our project, we want to investigate the trade-off between model performance and annotation costs. We use label ratio $r$ to represent annotation cost, so we need to make sure that the total fraction of labels needed in our experiment is consistent with the label ratio $r$. A simple example is shown in \cref{fig:sintel_pair}. Suppose we have a tiny training set of only two clean-final pairs $(c_1, f_1)$ and $(c_2, f_2)$, and we set the label ratio $r=0.5$. We have a total of four samples, so we need to select two of them to be labeled. If we sample clean and final images in pairs (as it is done in our experiments), the results may be like in \cref{fig:sintel_pair1}, where only $l_1$ (a half of the label set, consistent with $r=0.5$) is needed in the experiment. 

However, if we select clean and final samples separately, the selection may be like \cref{fig:sintel_pair2}. In this case, $c_1$ and $f_2$ are selected to be labeled, so both $l_1$ and $l_2$ are needed. This is inconsistent with $r=0.5$ because $100\%$ of the label set is needed, meaning that the annotation cost here is $100\%$. Therefore, the label ratio $r$ here does not represent the actual annotation cost needed in the experiment. Even though the labels are not used during $100\%$ of the training time (\eg, $l_1$ is not used when we train with $f_1$), we still need to pay full cost for annotating $l_1$ and $l_2$. Thus, this alternative sampling method does not support our investigation since $r$ does not reflect the annotation cost accurately.

Another option may be to use only one split (clean or final) for the experiments, \ie, training one semi-supervised model only on the clean split and another model only on the final split. This also solves the problem above that the label ratio $r$ does not reflect the true annotation cost. Nevertheless, this setting is largely different from most of the previous work, where both clean and final images are used to train one model that works on both passes at the same time. In this case, we are not able to compare with previous results. Such comparisons are crucial because we want to show that our semi-supervised models are significantly better than the state-of-the-art unsupervised models and also close to the supervised results.

\section{More Data and Results}\label{app:more_data_and_results}

\subsection{Raw Validation Data}\label{app:raw_validation_data}

Our raw data values are shown in \cref{tab:data_exp1,tab:data_exp2}. All pseudo error bars are obtained by taking the standard deviations in the last 50 epochs or 50k iterations. Semi-supervised training validation errors (Fig. 2 in the paper) are shown in \cref{tab:data_exp1}, and active learning validation errors (Fig. 3 in the paper) are shown in \cref{tab:data_exp2}.

\setlength{\tabcolsep}{5pt}
\begin{table}[t]
\centering
\small
\caption{Validation error for semi-supervised training on different datasets}
\label{tab:data_exp1}
\begin{tabular}{c|c|cc}
\hline
\multirow{2}{*}{{Label ratio} $r$} & {FlyingChairs}  & \multicolumn{2}{c}{{Sintel}}  \\
     & EPE/px        & clean EPE/px & final EPE/px  \\ \hline
0    & 3.066($\pm$0.044)  & 1.906($\pm$0.013)   & 2.933($\pm$0.010)    \\ \hline
0.05 & 2.369($\pm$0.033) & 1.850($\pm$0.014)   & 2.828($\pm$0.020)    \\
0.1  & 2.091($\pm$0.046)  & 1.776($\pm$0.018)   & 2.710($\pm$0.023)    \\
0.2  & 1.803($\pm$0.018)  & 1.691($\pm$0.011)   & 2.598($\pm$0.022)   \\
0.4  & 1.653($\pm$0.037) & 1.643($\pm$0.014)   & 2.349($\pm$0.031)    \\
0.6  & 1.560($\pm$0.039)  & 1.625($\pm$0.008)   & 2.281($\pm$0.022)    \\
0.8  & 1.550($\pm$0.043)  & 1.581($\pm$0.015)   & 2.281($\pm$0.015)   \\ \hline
1    & 1.439($\pm$0.052) & 1.651($\pm$0.018)   & 2.290($\pm$0.013)   \\ 
\hline \hline
\multirow{2}{*}{{Label ratio} $r$} & {FlyingThings3D} & \multicolumn{2}{c}{{KITTI}} \\
        & EPE/px     & 2012 Fl/\% & 2015 Fl/\% \\ \hline
0     & 12.037($\pm$0.500)    & 5.827($\pm$0.057) & 12.742($\pm$0.090) \\ \hline
0.05  & 10.588($\pm$0.444)    & 5.525($\pm$0.038) & 11.462($\pm$0.088) \\
0.1   & 10.205($\pm$0.694)    & 5.325($\pm$0.042) & 11.030($\pm$0.128) \\
0.2   & 9.584($\pm$0.132)    & 5.137($\pm$0.050) & 10.357($\pm$0.096) \\
0.4 & 8.395($\pm$0.307)  & 4.899($\pm$0.036) & 10.109($\pm$0.087) \\
0.6   & 8.296($\pm$0.154)    & 4.973($\pm$0.049) & 9.947($\pm$0.150) \\
0.8  & 7.833($\pm$0.152)   & 4.709($\pm$0.057) & 9.784($\pm$0.134) \\ \hline
1     & 7.876($\pm$0.283)   & 4.562($\pm$0.047) & 9.448($\pm$0.134 ) \\ \hline
\end{tabular}
\end{table}
\setlength{\tabcolsep}{1.4pt}

\setlength{\tabcolsep}{5pt}
\begin{table}[t]
\caption{Active learning validation errors, mean and std}
\label{tab:data_exp2}
\centering
\small
\begin{tabular}{c|c|cc}
\hline
\multirow{2}{*}{{Label ratio} $r$} & \multirow{2}{*}{{Method} }         &\multicolumn{2}{c}{{Sintel}} \\
 &  & clean EPE/px       & final EPE/px        \\ \hline
0           & -              & 1.906 ($\pm$0.013) & 2.933 ($\pm$0.010)  \\ \hline
            & random         & 1.850 ($\pm$0.014) & 2.828 ($\pm$0.020)   \\ \cline{2-4} 
0.05        & photo loss     & 1.807 ($\pm$0.010) & 2.731 ($\pm$0.015) \\
            & occ ratio      & \textbf{1.767 ($\bm\pm$0.019)} & \textbf{2.693 ($\bm\pm$0.017)}   \\
            & flow grad norm & 1.797 ($\pm$0.017) & 2.770 ($\pm$0.016)    \\ \hline
            & random         & 1.776 ($\pm$0.018) & 2.710 ($\pm$0.023)   \\ \cline{2-4} 
0.1         & photo loss     & 1.706 ($\pm$0.006) & 2.541 ($\pm$0.018)   \\
            & occ ratio      & \textbf{1.686 ($\bm\pm$0.013)} & \textbf{2.515 ($\bm\pm$0.018)}  \\
            & flow grad norm & 1.696 ($\pm$0.009) & 2.545 ($\pm$0.017)    \\ \hline
            & random         & 1.691 ($\pm$0.011) & 2.598 ($\pm$0.022)   \\ \cline{2-4} 
0.2         & photo loss     & 1.639 ($\pm$0.010)  & 2.383 ($\pm$0.025)   \\
            & occ ratio      & 1.643 ($\pm$0.013) & 2.373 ($\pm$0.018)   \\
            & flow grad norm & \textbf{1.631 ($\bm\pm$0.016)} & \textbf{2.299 ($\bm\pm$0.019)}  \\ \hline
1           & -              & 1.651 ($\pm$0.018) & 2.290 ($\pm$0.013)   \\ \hline \hline
\multirow{2}{*}{{Label ratio} $r$} & \multirow{2}{*}{{Method} }         & \multicolumn{2}{c}{{KITTI}} \\
 &      & 2012 Fl/\%   & 2015 Fl/\%     \\ \hline
0           & -                & 5.573 ($\pm$0.056) & 12.062 ($\pm$0.153) \\ \hline
            & random           & 5.363 ($\pm$0.080) & 11.456 ($\pm$0.158) \\ \cline{2-4} 
0.05        & photo loss       & 5.477 ($\pm$0.032) & 11.705 ($\pm$0.112)\\
            & occ ratio       & \textbf{5.256 ($\pm$0.040)} & \textbf{10.689 ($\pm$0.101)} \\
            & flow grad norm  & 5.353 ($\pm$0.047) & 10.994 ($\pm$0.171)  \\ \hline
            & random          & 5.273 ($\pm$0.034) & 10.480 ($\pm$0.108)  \\ \cline{2-4} 
0.1         & photo loss      & 5.175 ($\pm$0.040) & 10.441 ($\pm$0.087) \\
            & occ ratio        & 5.170 ($\pm$0.043) & \textbf{10.148 ($\pm$0.110)} \\
            & flow grad norm   & \textbf{5.159 ($\pm$0.039)} & 10.880 ($\pm$0.135)  \\ \hline
            & random          & 5.021 ($\pm$0.061) & 9.962 ($\pm$0.096) \\ \cline{2-4} 
0.2         & photo loss     & 4.934 ($\pm$0.033) & 9.759 ($\pm$0.140)  \\
            & occ ratio       & 4.929 ($\pm$0.043) & 9.736 ($\pm$0.147) \\
            & flow grad norm   & \textbf{4.837 ($\pm$0.046)} & \textbf{9.731 ($\pm$0.122)} \\ \hline
1           & -               & 4.446 ($\pm$0.034) & 8.545 ($\pm$0.086) \\ \hline
\end{tabular}
\end{table}
\setlength{\tabcolsep}{1.4pt}

\subsection{Benchmark Qualitative Results}\label{app:benchmark_qualitative_results}

\begin{figure}[t] 
    \centering
    \subfigure{
        \includegraphics[width=\linewidth]{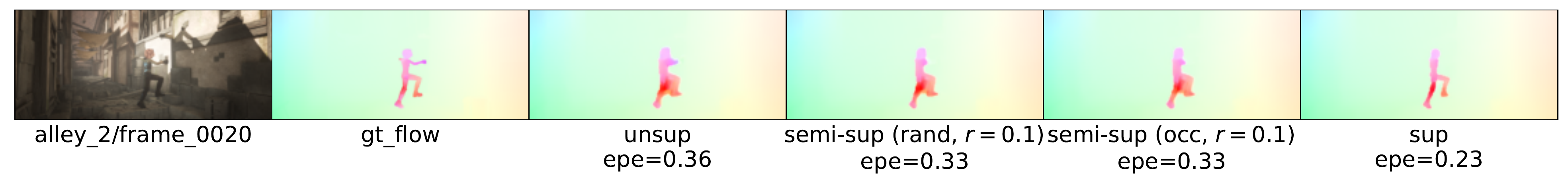} 
        \label{fig:qualitative_sintel_0}
    }
    \subfigure{
        \includegraphics[width=\linewidth]{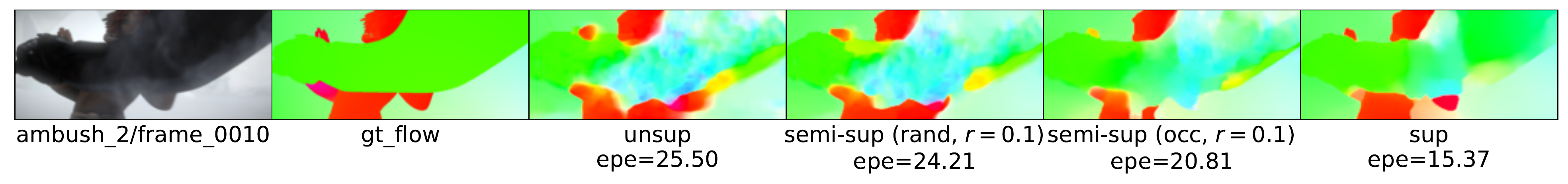} 
        \label{fig:qualitative_sintel_1}
    }
    \subfigure{
        \includegraphics[width=\linewidth]{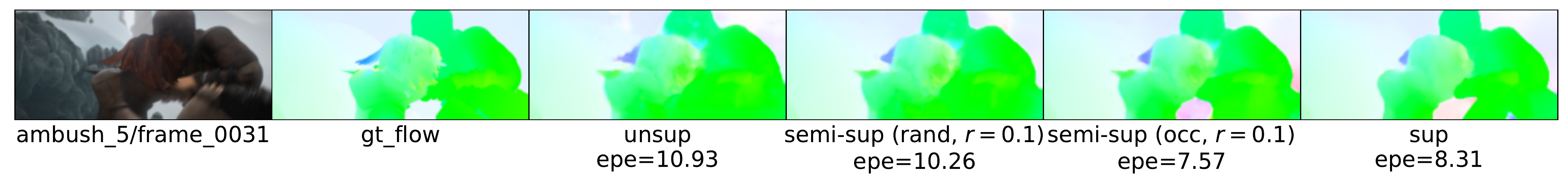} 
        \label{fig:qualitative_sintel_2}
    }
    \subfigure{
        \includegraphics[width=\linewidth]{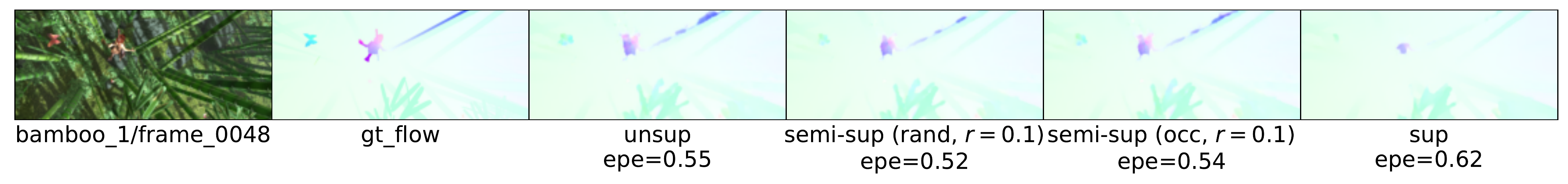} 
        \label{fig:qualitative_sintel_3}
    }
    \subfigure{
        \includegraphics[width=\linewidth]{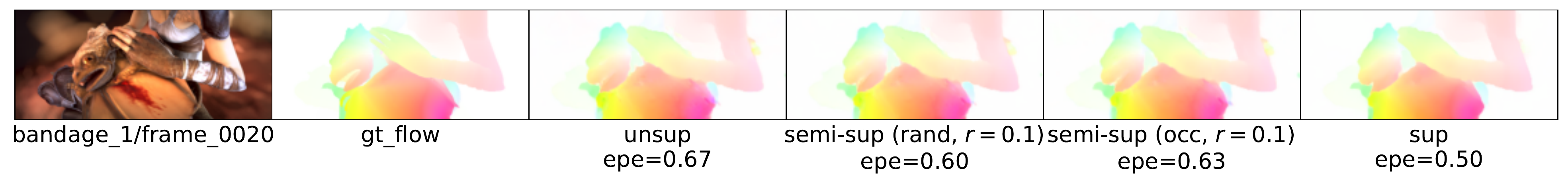} 
        \label{fig:qualitative_sintel_4}
    }
    \subfigure{
        \includegraphics[width=\linewidth]{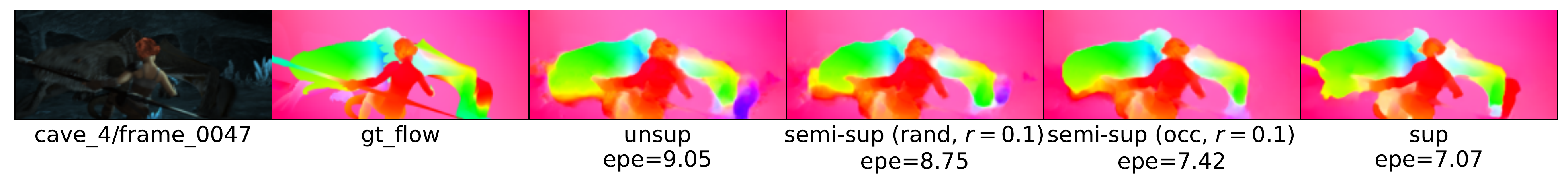} 
        \label{fig:qualitative_sintel_5}
    }
    \subfigure{
        \includegraphics[width=\linewidth]{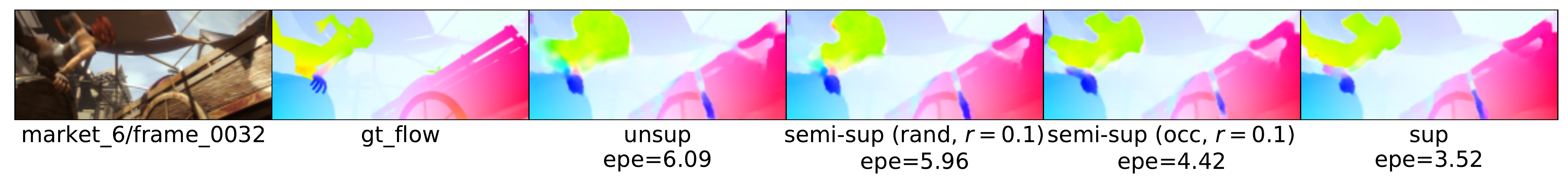} 
        \label{fig:qualitative_sintel_6}
    }
    \subfigure{
        \includegraphics[width=\linewidth]{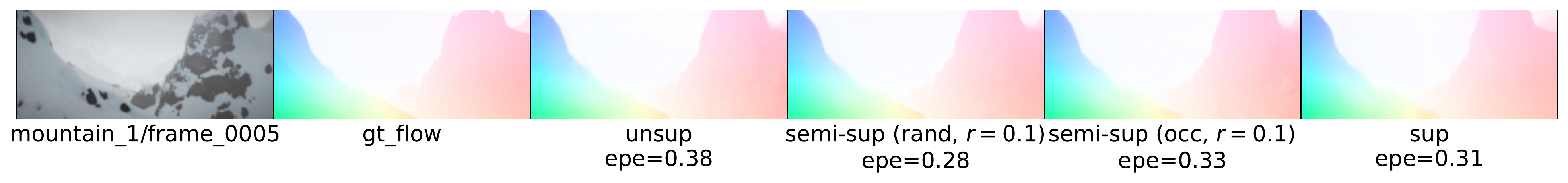} 
        \label{fig:qualitative_sintel_7}
    }
    \subfigure{
        \includegraphics[width=\linewidth]{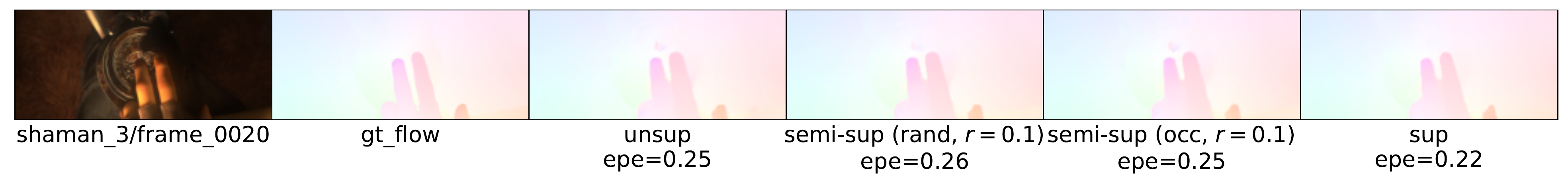} 
        \label{fig:qualitative_sintel_8}
    }
    \subfigure{
        \includegraphics[width=\linewidth]{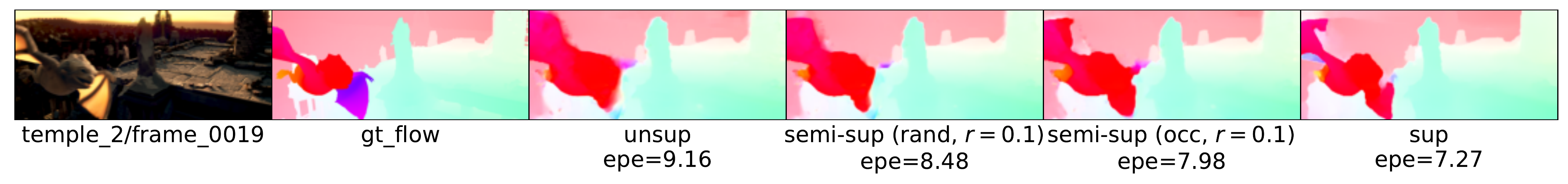} 
        \label{fig:qualitative_sintel_9}
    }
   \caption{Qualitative results on Sintel. Examples selected form our final pass validation split. Columns from left to right: the first frame image, the ground-truth flow, the unsupervised model prediction, the (random) semi-supervised model prediction (label ratio $r=0.1$), our active learning model prediction (label ratio $r=0.1$), the supervised model prediction. EPEs are shown in the subtitles. }
   \label{fig:qualitative_sintel}
\end{figure}
\begin{figure*}[t] 
    \centering
    \subfigure{
        \includegraphics[width=\linewidth]{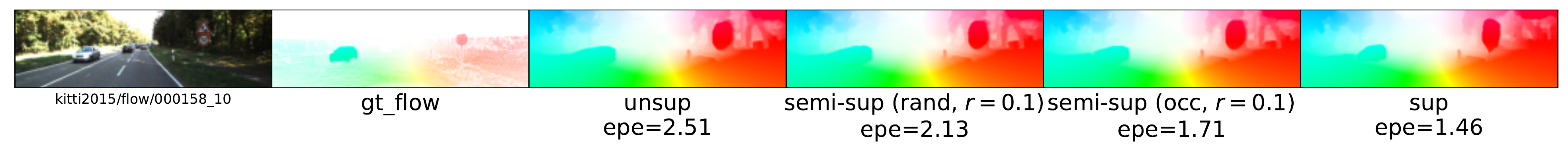} 
        \label{fig:qualitative_kitti_0}
    }
    \subfigure{
        \includegraphics[width=\linewidth]{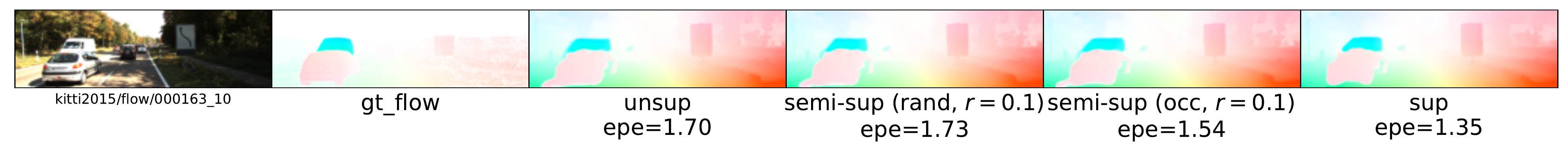} 
        \label{fig:qualitative_kitti_1}
    }
    \subfigure{
        \includegraphics[width=\linewidth]{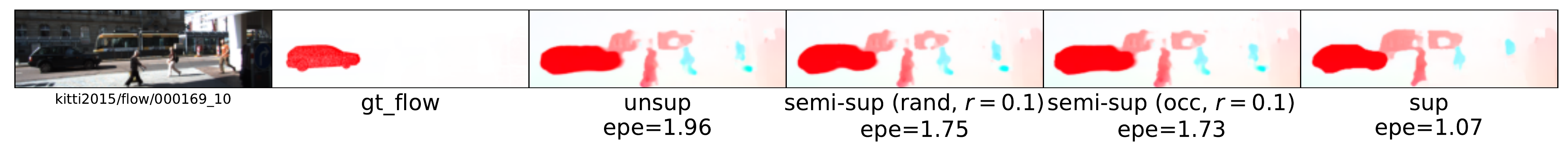} 
        \label{fig:qualitative_kitti_2}
    }
    \subfigure{
        \includegraphics[width=\linewidth]{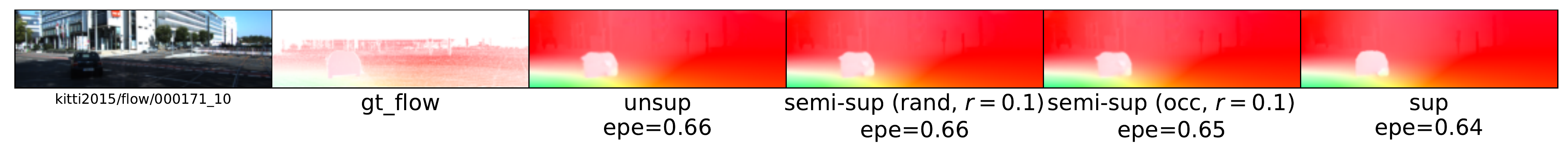} 
        \label{fig:qualitative_kitti_3}
    }
    \subfigure{
        \includegraphics[width=\linewidth]{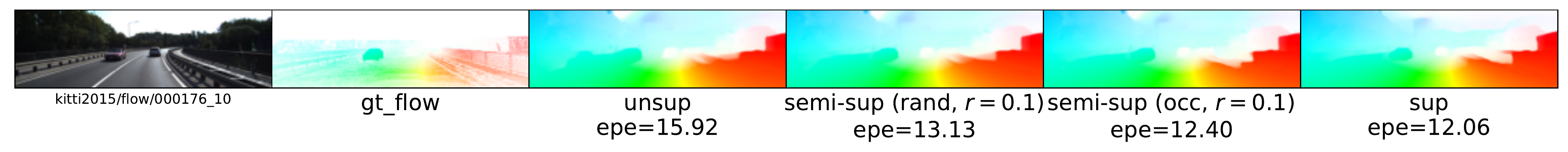} 
        \label{fig:qualitative_kitti_4}
    }
    \subfigure{
        \includegraphics[width=\linewidth]{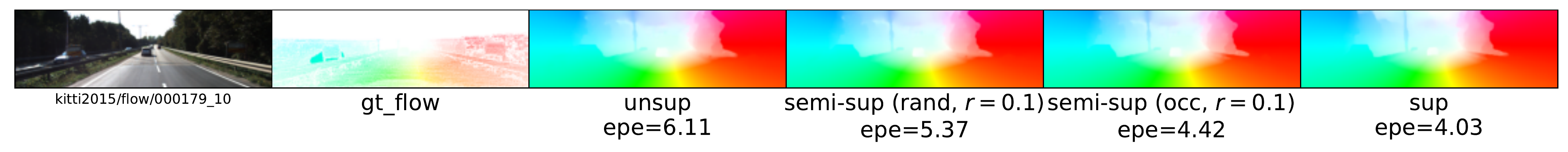} 
        \label{fig:qualitative_kitti_5}
    }
    \subfigure{
        \includegraphics[width=\linewidth]{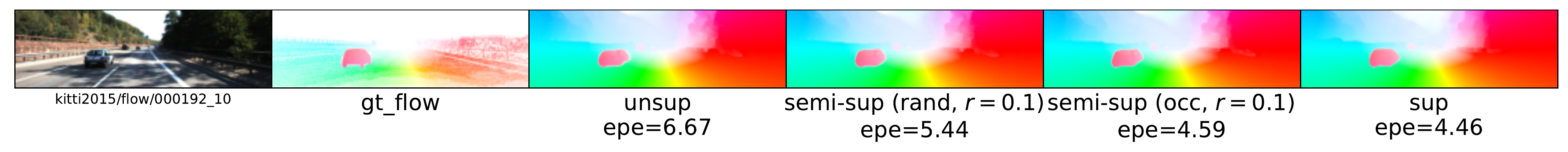} 
        \label{fig:qualitative_kitti_6}
    }
    \subfigure{
        \includegraphics[width=\linewidth]{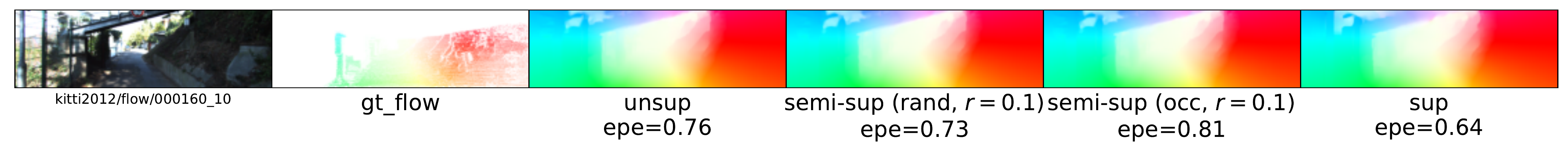} 
        \label{fig:qualitative_kitti_7}
    }
    \subfigure{
        \includegraphics[width=\linewidth]{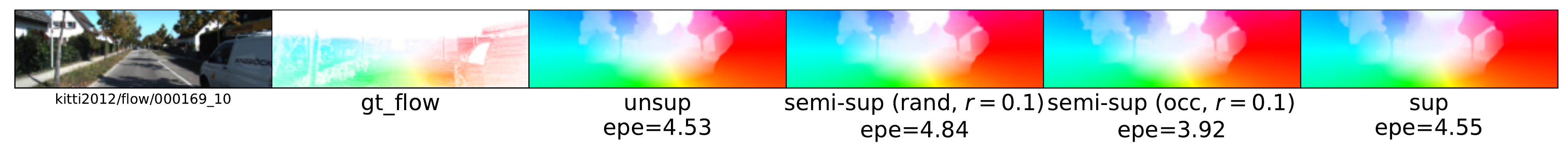} 
        \label{fig:qualitative_kitti_8}
    }
    \subfigure{
        \includegraphics[width=\linewidth]{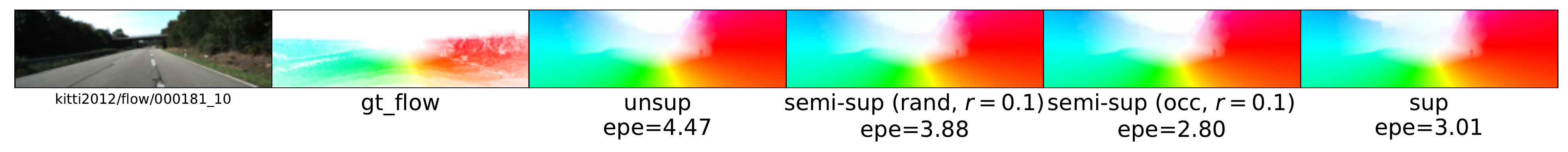} 
        \label{fig:qualitative_kitti_9}
    }
    \caption{Qualitative results on KITTI. Examples are selected from our validation split. Columns from left to right: the first frame image, the ground-truth flow, the unsupervised model prediction, the (random) semi-supervised model prediction (label ratio $r=0.1$), our active learning model prediction (label ratio $r=0.1$), the supervised model prediction. EPEs are shown in the subtitles. }
    \label{fig:qualitative_kitti}
\end{figure*}

Some qualitative results are shown in \cref{fig:qualitative_sintel}. We can see that our active learning method is especially effective at hard sequences like ``ambush'', ``cave'', ``market'' and ``temple'', and less effective at easy sequences where errors are already very small even for the unsupervised model. KITTI qualitative results are also shown in \cref{fig:qualitative_kitti}, where the differences are less visible with the naked eye.

\subsection{Analysis on More Uncertainty Scores}\label{app:analysis_on_more_uncertainty_scores}

We have also tried more metrics with heuristics defined as below.

\begin{itemize}
    \item \emph{flow norm}: the 2-norm of the estimated flow vectors averaged across the frame, used to reflect large motions.
    \item \emph{img grad norm}: the magnitude of gradients of the input images, used to reflect edges in the scene
    \item \emph{texture score}: used to evaluate whether the input images have good textures (high scores for good textures); computed based on Good Features to track~\cite{jianbo1994good}. We select 16*16 windows with stride=8. For each window, we compute the Z matrix for each pixel from image gradients and add them up to get a summed 2-by-2 positive semi-definite symmetric matrix. We compute the smaller eigenvalues (must positive) of the matrix for each window and take the average.
    \item \emph{color change}: used to indicate illumination change. We compute the color histogram for each RGB channel as well as its cumulative distribution. We compute the distances between the cumulative distributions of the first and second frames and take average across the RGB channels.
    \item \emph{param grad norm}: the norm of the loss gradients with respect to the network parameters. Intuitively, if a sample contributes large gradients to the network, it is likely that this sample does not fit well with the current network, so it may need labels.
    \item \emph{max corr vol}: the maximum value of the correlation volume at each pixel averaged across the whole frame. We use the correlation volume at the second-level decoder here. Intuitively, a large maximum correlation volume means that the pixel has a good match within the window, so the error may be small. 
\end{itemize}

\begin{figure}[htbp] 
    \centering
    \subfigure[Sintel]{
        \includegraphics[width=0.85\linewidth]{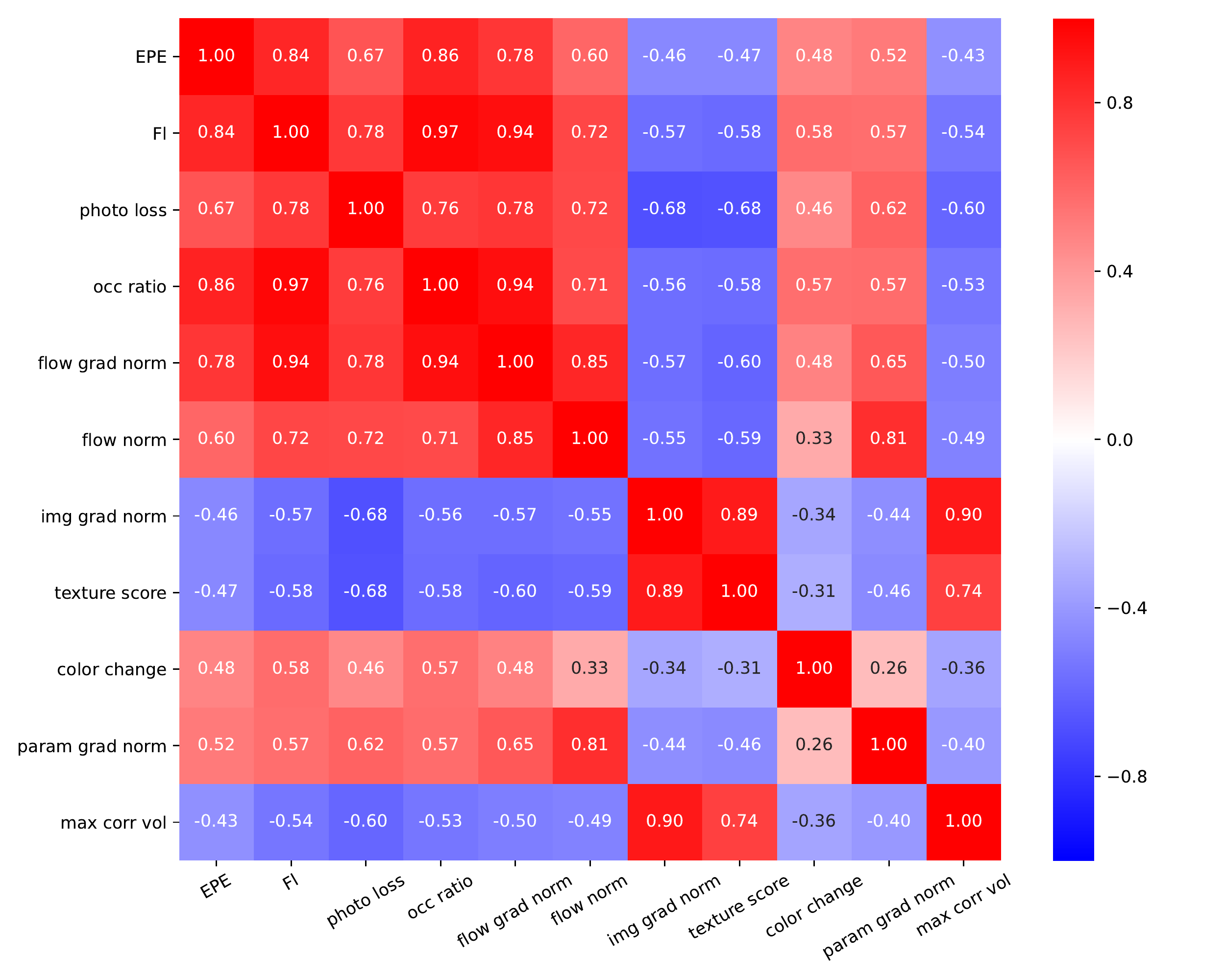} 
        \label{fig:sintel_corr_large}
    }
    \subfigure[KITTI]{
        \includegraphics[width=0.85\linewidth]{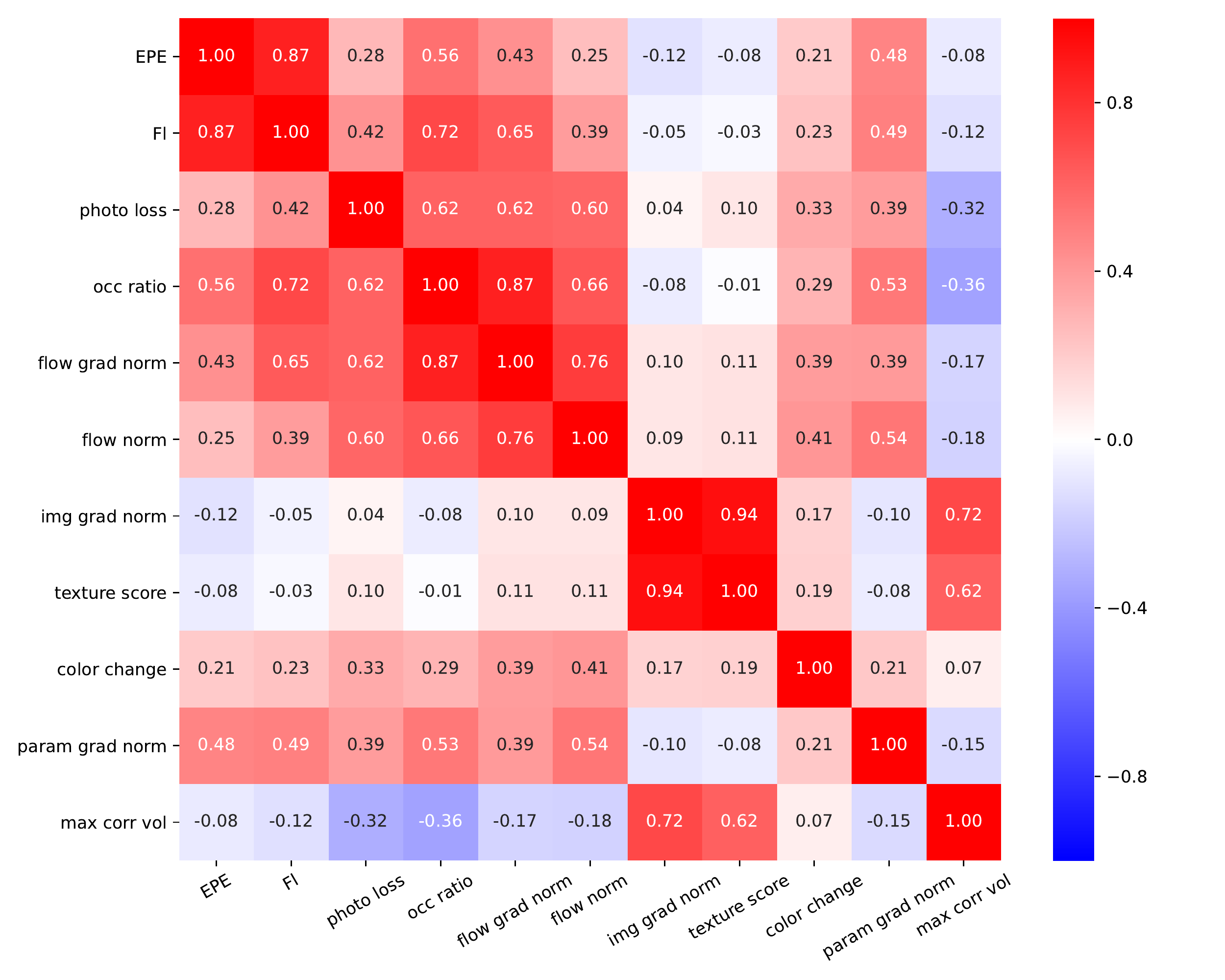} 
        \label{fig:kitti_corr_large}
    }
    \caption{The correlation matrices of more active learning criteria with sample errors}
    \label{fig:corr_large}
\end{figure}

We plot similar correlation matrices (as the ones in the last part of the main paper) with all our metrics in \cref{fig:corr_large}. We can see that all metrics are more or less consistent with our intuitions. Note that the ``img grad norm'' and ``texture score'' are negatively correlated with the errors because larger values indicate better textures and thus smaller estimation errors. Also, ``max corr vol'' is negatively correlated because larger values indicate better matches found for the first image pixels. From \cref{fig:corr_large}, we can see that the metrics that are more correlated with the errors are ``occ ratio'', ``flow grad norm'', ``photo loss'', which are then used in our experiments.

Comparing the Sintel correlation matrix (\cref{fig:sintel_corr_large}) from that of KITTI (\cref{fig:kitti_corr_large}), we can see that the Sintel metrics are generally more correlated with the errors, whereas KITTI metrics are generally less effective in detecting the samples of large errors. Especially for the texture related scores like ``img grad norm'' and ``texture score'', Sintel errors have correlations around 0.5, but KITTI errors are almost independent of the sample errors. We guess it may be because KITTI can already achieve pretty decent results by merely learning the flow distribution patterns (the looming motion), so it does not have to track every patch closely.

\clearpage

\end{document}